\documentclass[12pt]{iopart}

\usepackage[numbers,square,sort&compress]{natbib}

\expandafter\let\csname equation*\endcsname\relax
\expandafter\let\csname endequation*\endcsname\relax
\usepackage{amsmath}
\usepackage[T1]{fontenc}
\renewcommand*\vec[1]{\boldsymbol{\mathbf{#1}}}
\usepackage{placeins}
\usepackage{mathtools}
\usepackage{hyperref}
\usepackage[capitalize,noabbrev]{cleveref}
\crefname{appendix}{}{}
\usepackage{amssymb}
\usepackage{amsthm}
\usepackage{enumitem}
\usepackage{xcolor}

\theoremstyle{plain}
\newtheorem{theorem}{Theorem}[section]

\newtheorem{corollary}[theorem]{Corollary}
\theoremstyle{definition}

\theoremstyle{remark}

\begin{document}

\title[Anti-Correlated Noise in Epoch-Based Stochastic Gradient Descent]{Anti-Correlated Noise in Epoch-Based Stochastic Gradient Descent: Implications for Weight Variances in Flat Directions}

\author{Marcel K\"uhn$^{1}$ and Bernd Rosenow$^{1}$}

\address{$^{1}$Institute for Theoretical Physics, University of Leipzig, 04103 Leipzig, Germany}
\vspace{10pt}
\begin{indented}
\item[]8 October 2025
\end{indented}

\begin{abstract}
Stochastic Gradient Descent (SGD) has become a cornerstone of neural network optimization due to its computational efficiency and generalization capabilities. However, the gradient noise introduced by SGD is often assumed to be uncorrelated over time, despite the common practice of epoch-based training where data is sampled without replacement. In this work, we challenge this assumption and investigate the effects of epoch-based noise correlations on the stationary distribution of discrete-time SGD with momentum. Our main contributions are twofold: First, we calculate the exact autocorrelation of the noise during epoch-based training under the assumption that the noise is independent of small fluctuations in the weight vector, revealing that SGD noise is inherently anti-correlated over time. Second, we explore the influence of these anti-correlations on the variance of weight fluctuations. 
We find that for directions with curvature of the loss greater than a hyperparameter-dependent crossover value, the conventional predictions of isotropic weight variance under stationarity, based on uncorrelated and curvature-proportional noise, are recovered. Anti-correlations have negligible effect here.
However, for relatively flat directions, the weight variance is significantly reduced, leading to a considerable decrease in loss fluctuations compared to the constant weight variance assumption. Furthermore, we present a numerical experiment where training with these anti-correlations enhances test performance, suggesting that the inherent noise structure induced by epoch-based training may play a role in finding flatter minima that generalize better.
\end{abstract}

\vspace{2pc}
\noindent{\it Keywords}: Stochastic Gradient Descent, Asymptotic Analysis, Discrete Time, Hessian
%\maketitle

\section{Introduction}

Initially developed to address the challenges of computational efficiency in neural networks, stochastic gradient descent (SGD) has exhibited exceptional effectiveness in managing large datasets compared to the full gradient methods \cite{Bottou.1991}. It has since garnered widespread acclaim in the machine learning domain \cite{LeCun.2015}, with applications spanning image recognition \cite{Krizhevsky.2012, Simonyan.2015, He.2016}, natural language processing \cite{Sutskever.2014, Brown.2020}, and mastering complex games beyond human capabilities \cite{Silver.2017}. Alongside its numerous variants \cite{Duchi.2011, Kingma.2015, Schmidt.2021}, SGD remains the cornerstone of neural network optimization.

SGD's success can be attributed to several key properties, such as rapid escape from saddle points \cite{Ge.2015} and its capacity to circumvent "bad" local minima, instead locating broad minima that generally lead to superior generalization \cite{Hochreiter.1997, Keskar.2017, Jastrzebski.2017, Smith.2018, Xie.2021, Wojtowytsch.2021}. This is often ascribed to anisotropic gradient noise \cite{Hoffer.2017, Sagun.2018, Zhang.2018, Zhang.2019, Zhu.2019, Li.2020, Ziyin.2022}. Nonetheless, recent empirical research posits that even full gradient descent, with minor adjustments, can achieve generalization performance comparable to that of SGD \cite{Geiping.2022}.

To deepen our understanding of neural network training dynamics, several studies have investigated the limiting behavior of network weights during the later stages of training \cite{Yaida.2019, Kunin.2021}. Of particular interest is the behavior of weight fluctuations near a minimum of the training loss function \cite{Mandt.2017, Jastrzebski.2017, Liu.2021}. 
The variance of these fluctuations reveals how strongly different directions in parameter space are explored and has been used, for example, to quantify escaping efficiency from different minima \cite{Zhu.2019, Liu.2021}, offering valuable insights into optimization behavior.
Empirical evidence suggests that the covariance matrix $\mathbf{C}$ associated with SGD noise is proportional to the Hessian matrix $\mathbf{H}$ of the training loss function \cite{Sagun.2018, Zhang.2018, Zhang.2019, Zhu.2019, Thomas.2020, Xie.2021}, although this proportionality may not hold under certain conditions \cite{Ziyin.2022}. Here, $\mathbf{C}$ is defined as the covariance of the difference between the minibatch gradient and the full-batch gradient.
Consequently, theoretical analyses predict that the stationary covariance matrix of the weights, $\mathbf{\Sigma}$, becomes isotropic when the learning rate is sufficiently small \cite{Jastrzebski.2017, Liu.2021, Kunin.2021}. However, recent empirical studies \cite{Feng.2021} have identified significant anisotropy in $\mathbf{\Sigma}$.

In this work, we present both theoretical and empirical analyses of weight fluctuations during the later stages of training, accounting for the emergence of anti-correlations in the noise produced by SGD, which stem from the prevalent epoch-based learning schedule. As a result of these anti-correlations, we discover that the covariance matrix $\bf \Sigma$  displays anisotropy and is smaller than expected in a subspace of weight directions corresponding to Hessian eigenvectors with small eigenvalues, while maintaining the isotropy of $\bf \Sigma$ in directions associated with Hessian eigenvectors possessing large eigenvalues. Our theoretical predictions are validated through the analysis of a neural network's training within a subspace of its top Hessian eigenvectors.

In addition, we demonstrate that for a small convolutional network trained on CIFAR10 the anti-correlations in SGD noise described above significantly increase the test accuracy. By linking this result to a previous study on artificially added anti-correlated noise and its benefits \cite{Orvieto.2022}, we argue that the anti-correlations in SGD noise suppress diffusion in flat directions, and in this way may contribute to finding flatter minima with better test accuracy.

\subsection*{Our contributions}
\begin{itemize}

\item We uncover generic anti-correlations in SGD noise that result from the common practice of drawing training examples without replacement. We calculate the autocorrelation function of SGD noise under the assumption that the noise is independent of small fluctuations in the weight vector over time. The anti-correlations arise because the noise sums to zero over an epoch where each example is presented once.

\item Using the computed autocorrelation function, we develop a theory that elucidates the relationship between variances along different Hessian eigenvectors. The weight space partitions into two groups based on Hessian eigenvalues: for eigenvectors with eigenvalues exceeding a hyperparameter-dependent crossover value $\lambda_{\textrm{cross}}$, the isotropic variance prediction holds; for those with smaller eigenvalues, anti-correlations reduce the variance, resulting in values proportional to the eigenvalues and smaller than the isotropic prediction. This theory accounts for the discrete nature of SGD, avoids continuous-time approximations, and incorporates heavy-ball momentum.

\item For each Hessian eigenvector, there exists an intrinsic correlation time of  update steps, introduced by the loss landscape and proportional to $1/\lambda_i$, where $\lambda_i$ is the corresponding Hessian eigenvalue. Additionally, the noise anti-correlations give rise to a direction-independent noise correlation time $\tau_\textrm{SGD}$ on the order of one epoch. The smaller of these two timescales determines the actual correlation time $\tau_i$ for the update steps along a given eigenvector. Since weight variances are proportional to $\tau_i$, this leads to the emergence of two distinct variance-curvature relationships.

\item We validate our theoretical predictions by analyzing the training of a neural network within a subspace spanned by its top Hessian eigenvectors. Utilizing Hessian eigenvectors offers advantages over principal component analysis  of the weight trajectory, as used in a previous empirical study \cite{Feng.2021}, where finite-size effects introduced significant artifacts.

\item We observe that drawing examples without replacement during training leads to higher test accuracy compared to drawing with replacement. By linking our findings to a prior study \cite{Orvieto.2022} on artificially added anti-correlated noise, we propose that the higher test accuracy is connected to the anti-correlated noise inherent in the without-replacement sampling, which may encourage convergence to flatter minima. This interpretation is further supported by our analysis of escaping efficiency, where we show that reduced variance in flat directions due to noise anti-correlations potentially stabilizes optimization in low-curvature regions and enhances generalization. In particular, for the distribution of the 5,000 largest Hessian eigenvalues derived from our LeNet case study, anti-correlations affect the escaping efficiency significantly. We observe a 62\% reduction in escaping efficiency for this particular minimum with many flat directions, compared to a setup without anti-correlated noise.

\end{itemize}

\section{Background}

We consider a neural network characterized by its weight vector, $\vec{\theta} \in \mathbb{R}^d$. The network is trained on a set of $N$ training examples, each denoted by $x_n$, where $n$ ranges from 1 to $N$. The training  loss function, defined as ${L(\vec{\theta}) \coloneqq \frac{1}{N} \sum_{n=1}^N l(\vec{\theta}, x_n)}$, represents the average of individual losses incurred for each training example $l(\vec{\theta}, x_n)$.

It is common practice to add some kind of momentum to the SGD algorithm when training a network as it leads to faster convergence, has a smoothing effect, and helps in escaping local minima.  Therefore, to keep the analysis general, we consider a training process that employs stochastic gradient descent augmented with heavy-ball momentum. This approach updates the network parameters according to the following rules:
\begin{align}
 \vec{g}_k(\vec{\theta}) = \frac{1}{S} \sum_{n \in \mathcal{B}_k} \vec{\nabla} l(\vec{\theta}, x_n) \ , \qquad \vec{v}_{k} =  -\eta \vec{g}_{k}(\vec{\theta}_{k-1}) + \beta \vec{v}_{k-1} \ , \qquad \vec{\theta}_{k} = \vec{\theta}_{k-1} + \vec{v}_{k} \ .
\end{align}
Here, $k$ signifies the discrete update step index, $\eta$ is the learning rate, and $\beta$ is the momentum parameter. The stochastic gradient at each step is computed with respect to a batch of $S \ll N$ random examples. Each batch is denoted by  $\mathcal{B}_k = \{n_1, \dots, n_S\}$, where $n_j \in \{1, \dots, N\}$.
The training process is structured into epochs. During each epoch, every training example is used exactly once,  implying that the examples are drawn without replacement and do not recur within the same epoch.

In the realm of SGD as opposed to full gradient descent, we introduce noise, denoted as $\Vec{\delta g}_k(\Vec{\theta}) \coloneqq \Vec{g}_k(\Vec{\theta}) - \Vec{\nabla}L(\Vec{\theta})$, with a   covariance matrix  $\vec{C}(\Vec{\theta}) \coloneqq \textrm{cov}\bigl(\Vec{\delta g}_k(\Vec{\theta}), \Vec{\delta g}_k (\Vec{\theta})\bigr)$. The noise covariance matrix is known to be proportional to the gradient sample covariance matrix $ \Vec{C_0}(\Vec{\theta}) \coloneqq \frac{1}{N-1} \sum_{n=1}^N \Vec{\nabla}\left[ l(\vec{\theta}, x_n) - L(\Vec{\theta}) \right]\Vec{\nabla}^\top\left[ l(\vec{\theta}, x_n) - L(\Vec{\theta}) \right]$. A detailed derivation of this relation, particularly for the case of drawing without replacement, is provided in \cref{sec:appendix_corr_calc}.

As in previous studies, our primary theoretical focus is on the asymptotic -- or limiting -- covariance matrix of the weights, denoted by $\vec{\Sigma} \coloneqq \textrm{cov}\bigl(\Vec{\theta}_k, \Vec{\theta}_k\bigr)$. This means we are interested in the covariance computed over an infinite run of SGD optimization, rather than the covariance at a fixed update step $k$ computed over multiple runs of SGD (see \cref{sec:appendix_limit_definition}).

To better understand the behavior of the weight variances, we further examine the covariance matrix of the velocities, $\vec{\Sigma_v} \coloneqq \textrm{cov}\bigl(\Vec{v}_k, \Vec{v}_k\bigr)$. We then explore the ratio of the weight variance to the velocity variance in any given direction, which we denote as $\tau_i$. Under general assumptions, this ratio equates to the velocity correlation time of the corresponding direction (see \cref{sec:theory_corr_time}).

\section{Related Work}

\subsection{Hessian and gradient sample covariance}
\label{sec:related_works_hessian_noise}

The assumption of a strong alignment between the gradient sample covariance $\vec{C_0}$ and the Hessian matrix of the training loss function $\vec{H}$ is widely used in the literature \cite{Jastrzebski.2017, Zhang.2019, Liu.2021}. Various theoretical arguments suggest that when a neural network's output closely matches the example labels, these two matrices should be similar \cite{Jastrzebski.2017, Sagun.2018, Zhang.2018, Zhu.2019, Martens.2014}. However, as highlighted by Thomas et al.~\cite{Thomas.2020}, even slight deviations between network predictions and labels can theoretically disrupt this relationship. Nevertheless, empirical observations  often reveal a strong alignment between the gradient sample covariance and the Hessian matrix near a minimum.

Zhang et al.~\cite{Zhang.2019} investigated this assumption numerically. They analyzed both matrices in a specific basis that exhibits varying curvature across different directions. Their findings showed a close correspondence between curvature and gradient variance along a given direction in a convolutional image recognition network, and a reasonably good relationship in a transformer model.
Similarly, Thomas et al.~\cite{Thomas.2020} provided both theoretical arguments and empirical evidence across different image recognition architectures. While they did not find an exact match between the two matrices, they observed a proportionality indicated by a high cosine similarity.
Xie et al.~\cite{Xie.2021} also explored this relationship using an image recognition network. In the eigenspace of the Hessian matrix, they plotted entries within a specific interval against corresponding entries from the gradient sample covariance and observed a close match.

Ziyin et al.~\cite{Ziyin.2022} emphasized the importance of considering the conditions under which the approximation of proportionality between the two matrices is valid, noting that the approximation may be questionable otherwise. However, the conditions they specify -- particularly being near a minimum with low loss -- are consistent with our empirical setup (see \cref{sec:analysis_setup} and \cref{sec:appendix_loss_evolution}).

\subsection{Limiting Dynamics and Weight Fluctuations}

Several studies have examined the limiting dynamics of SGD, often modeling it as a stochastic differential equation (SDE). Researchers such as  Mandt et al.~\cite{Mandt.2017} and Jastrz\k{e}bski et al.~\cite{Jastrzebski.2017} commonly approximate the loss near a minimum as a quadratic function, representing the SDE as a multivariate Ornstein-Uhlenbeck (OU) process. This process suggests a stationary weight distribution with Gaussian fluctuations. Jastrz\k{e}bski et al.~\cite{Jastrzebski.2017} further assume that the gradient covariance is proportional to the Hessian, observing under these conditions that the weight fluctuations are isotropic. Kunin et al.~\cite{Kunin.2021}, who also incorporate momentum into their analysis, predict and empirically verify isotropic weight fluctuations. Chaudhari \& Soatto~\cite{Chaudhari.2018} investigate the SDE without assuming a quadratic loss or equilibrium, gaining insights via the Fokker-Planck equation.

Alternatively, some studies derive relationships from a stationarity assumption rather than a continuous-time approximation \cite{Yaida.2019, Liu.2021, Ziyin.2022}. Yaida~\cite{Yaida.2019} assumes that the weight trajectory follows a stationary distribution and derives general fluctuation-dissipation relations from this premise. Liu et al.~\cite{Liu.2021} go further by assuming a quadratic loss function, enabling them to derive exact relations for the weight variance of SGD with momentum. If the gradient covariance is additionally assumed to be proportional to the Hessian, their results also predict that the weight variance is approximately isotropic, except in directions where the product of the learning rate and the Hessian eigenvalue is significantly large.

These theoretically computed weight variances are explicitly applied in various contexts, such as calculating the escape rate from a minimum or assessing the approximation error in SGD -- which captures the additional training error attributed to noise \cite{Liu.2021}.

Feng \& Tu~\cite{Feng.2021} present a phenomenological theory based on their empirical findings, which, unlike Kunin et al.~\cite{Kunin.2021}, also accounts for flat directions. They describe a general inverse variance-flatness relationship by analyzing the weight trajectories of different image recognition networks via principal component analysis. They discovered a power law relationship between the curvature of the loss and the weight variance $\sigma_{\theta, i}^2$ in any given direction, where higher curvature corresponds to higher variance. They also observed that both the velocity variance $\sigma_{v, i}^2$ and the correlation time $\tau_i$ are larger for higher curvatures.

In our approach, we avoid the continuous-time approximation and instead base our results on the assumption that the weights adhere to a stationary distribution near a quadratic minimum.

\section{Theory}
\subsection{Autocorrelation of the noise}
\label{sec:correlation_main}

\begin{figure}[ht]
    \centering
    \includegraphics[width=0.8\linewidth]{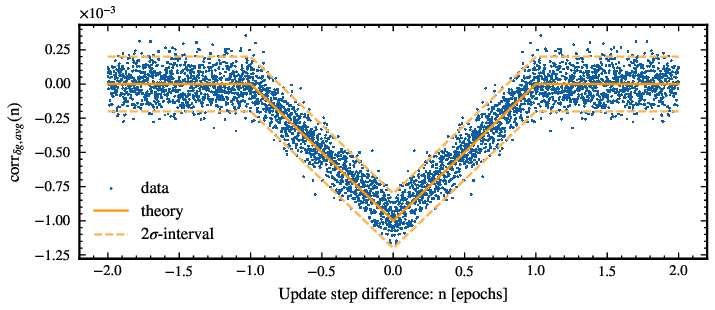}
    \caption{Autocorrelations of the SGD noise observed over a span of 20 epochs, equivalent to 20,000  update steps. This data is collected from a later phase in the training process. The autocorrelation is projected onto 5,000  Hessian eigenvectors, and the result is averaged. The theoretical prediction \cref{eq:noise_corr} is also displayed along with a $2\sigma$-interval, where $\sigma$ represents the expected standard deviation of the SGD noise. The zero-point correlation is omitted as it is inherently equal to one.}
    \label{fig:noise_corr}
\end{figure}

We consider the correlation between two noise terms arising from different SGD update steps. Specifically, when examples are sampled without replacement while keeping the weight vector $\vec{\theta}$ constant, inherent anti-correlations emerge in the noise. Below, we provide a conceptual motivation for this phenomenon and refer the reader to \cref{sec:appendix_corr_calc} for a detailed derivation.

Suppose the ratio $M \coloneqq N/S$ is an integer, where $N$ is the total number of examples and $S$ is the batch size. This ensures that each epoch consists of $M$ batches, with every example presented exactly once per epoch. For a fixed weight vector $\vec{\theta}$, the mean of the gradients computed over one epoch, $\vec{g}_k(\vec{\theta})$, equals the total gradient $\vec{\nabla}L(\vec{\theta})$. Consequently, the sum of the noise components $\vec{\delta g}_k(\vec{\theta}) = \vec{g}_k(\vec{\theta}) - \vec{\nabla}L(\vec{\theta})$ over one epoch must be zero.  This implies that if a noise term points in a particular direction at the beginning of an epoch, subsequent noise terms within the same epoch are constrained to partially cancel it out, leading to anti-correlations.

Therefore, the anti-correlation between two noise terms within the same epoch can be expressed as a negative correlation proportional to the equal-time correlation, scaled by a factor of $\frac{1}{M-1}$, where $M$ is the number of batches per epoch.
We assume the order of the examples within each new epoch is reshuffled, so any two noise terms from different epochs are uncorrelated.
Next, we consider the probability that two batches $k$ and $k+h$, separated by $h$ update steps, belong to the same epoch. This probability is given by $\frac{M - |h|}{M}$ for $|h| \leq M$, and zero otherwise. Combining this with the anti-correlation derived above, we arrive at the correlation formula:

\medskip
\begin{theorem}
If the total number of examples $N$ is an integer multiple of the batch size $S$ and the parameters $\vec{\theta}$ of a network are kept fixed, then the autocorrelation formula for the gradient noise of an epoch-based learning schedule, where the examples for each new epoch are drawn without replacement, is given by
\begin{align}
    {\rm cov}\bigl(\Vec{\delta g}_k(\Vec{\theta}), \Vec{\delta g}_{k+h}(\Vec{\theta})\bigr) =  \begin{cases}
        \vec{C}(\Vec{\theta}) \, , & h=0\\
        -\frac{M - |h|}{M(M-1)}\vec{C}(\Vec{\theta}) \, ,&1\leq|h|\leq M \\
        0 \, , &|h|>M
    \end{cases} \ , \label{eq:noise_corr}
\end{align}
where $M \coloneqq N/S$ signifies the number of batches per epoch.
\end{theorem}
\smallskip

The actual noise autocorrelation is illustrated in \cref{fig:noise_corr} and confirms the V-shaped theoretical autocorrelation structure, with the experimental details elaborated in \cref{sec:num_corr}. The complete calculation is available in \cref{sec:appendix_corr_calc}, highlighting the relationship $\vec{C}(\Vec{\theta}) = (1/S)\cdot(1-S/N)\cdot\vec{C_0}(\Vec{\theta})$ with the gradient sample covariance matrix $\vec{C_0}$.

Strictly speaking, the formula derived above applies only to a static weight vector. During training, the weights change with each update step, potentially altering this relationship.
A theoretical extension accounting for fluctuating weights around a minimum is presented in \cref{sec:appendix_non_static_weights}, demonstrating that the predicted anti-correlations still hold approximately under realistic training dynamics in a late stage of training.
Empirically, we find that in the later stages of training, the theoretical prediction given by \cref{eq:noise_corr} closely matches the actual training dynamics, as evidenced by the strong agreement with the data shown in \cref{fig:noise_corr}. Further numerical investigations of the LeNet network reveal, that even at the beginning of training, when the network weights are clearly changing significantly, SGD noise still exhibits anti-correlations in good agreement with our theoretical prediction (see \cref{sec:appendix_correlation_start}), although our analytic theory relates only to a late phase of training.

When we sample the examples with replacement during training, there are no anti-correlations (see \cref{sec:appendix_with_replacement}). 
And while the above result assumes that $M = N/S$ is an integer, we find that the theoretical predictions remain approximately valid even when this condition is not exactly satisfied. 
In this case, the noise terms do not exactly add up to zero, but the variance of this sum over one epoch is reduced by a factor of $1/M$ compared to sampling with replacement.
A detailed discussion of the residual noise and its effect in the non-integer case is provided in \cref{sec:appendix_non_integer_M}. 
Further numerical investigations show that in the case where $N$ is not an integer multiple of $S$, the later derived relations for the variances still hold (see \cref{sec:appendix_hyperparameters}).

\subsection{Understanding Variance Structure via Velocity Variance and Correlation Time}
\label{sec:theory_corr_time}

To gain a more comprehensive understanding of the weight variance behavior, we separate it into two interpretable components: the velocity variance and a correlation time. Here, we define the velocity $\vec{v}_k$ as the parameter update at step $k$, i.e., $\vec{v}_k = \vec{\theta}_{k+1} - \vec{\theta}_k$. Focusing on the velocity helps to isolate how stochasticity and curvature interact dynamically. We label the ratio between weight and velocity variance, scaled by a factor of two, as the correlation time $\tau_i$ (see \cref{eq:corr_time_def}). This definition aligns with that of the velocity correlation time, hence justifying the nomenclature. The equivalence stands under general assumptions, such as a sufficiently fast decay of the weight correlations, $\textrm{cov}\bigr(\vec{\theta}_{k}, \vec{\theta}_{k+n}\bigr)$, which are satisfied in our problem setup described in \cref{sec:var_main} (see \cref{sec:appendix_var_tau_assumptions}), where the weight correlations will even decay exponentially fast.

To motivate this decomposition, consider the simple one-dimensional model of SGD with replacement in a quadratic loss ($L(\theta) = \frac{1}{2}\lambda\theta^2$) without momentum: ${\theta_{k+1} = (1 - \eta \lambda)\theta_k + \eta \delta g_k}$, where $\delta g_k$ denotes uncorrelated zero-mean gaussian gradient noise for simplicity. Assuming stationarity, it is straightforward to solve for the weight variances. For $\eta \lambda \ll 1$, this yields $\sigma_\theta^2=\frac{\eta}{2\lambda}\sigma_{\delta g}^2$ and exponentially decaying correlations $\langle \theta_{k+h} \theta_k \rangle = \sigma_\theta^2 e^{-h/\tau}$ with $\tau = \frac{1}{\eta\lambda}$, since $(1-\eta\lambda)^h \approx e^{-\eta\lambda h}$. The velocities then exhibit anti-correlations on the same timescale, $\langle v_{k+h} v_k \rangle = -\frac{1}{\tau^2} e^{-h/\tau} \sigma_\theta^2 \quad (|h| \geq 1) $. Additionally, it follows that $\sigma_v^2 = \eta^2 \sigma_{\delta g}^2$, and similarly to the theorem below, one recovers $\sigma_\theta^2 = \frac{1}{2} \tau \sigma_v^2$. This illustrates that the velocity variance is determined by the noise, while the correlation time $\tau$ encapsulates the influence of the loss landscape curvature. For example, if the gradient noise variance scales with curvature, $\sigma_{\delta g}^2 \propto \lambda$, the model yields constant weight variance independent of $\lambda$ -- the known prediction in the with-replacement case. In the more complex case of sampling without replacement, anti-correlations in the noise modify this picture and induce a curvature-dependent suppression of $\tau$, giving rise to richer variance behavior, as explored in the next section. 

Nonetheless, the equivalence between the velocity correlation time and the ratio of weight and velocity variance is a more general result. The assumptions for the following theorem encompass: (i) Existence and finiteness of $\vec{\Sigma} \coloneqq $ cov$(\vec{\theta}_k, \vec{\theta}_k)$, $\vec{\Sigma_v} \coloneqq $ cov$(\vec{v}_k, \vec{v}_k)$, and $\langle \vec{\theta} \rangle$. (ii) $\lim_{n \to \infty}\textrm{cov}\bigr(\vec{\theta}_{k}, \vec{\theta}_{k+n}\bigr) = 0$. (iii) $\lim_{n \to \infty}n\cdot\textrm{cov}\bigr(\vec{\theta}_{k}, \vec{\theta}_{k+n} - \vec{\theta}_{k+n+1}\bigr) = 0$. For example, the latter two assumptions hold true if the weight correlation function decays as $\textrm{cov}\bigr(\vec{\theta}_{k}, \vec{\theta}_{k+n}\bigr) \propto n^{-2}$ or faster.

\medskip
\begin{theorem}
\label{thm:correlation_time}
Under the mentioned assumptions, which are satisfied by the problem setup described in \cref{sec:var_main}, it holds that
\begin{align}
 \tau_i \coloneqq \frac{2 \sigma_{\theta, i}^2}{\sigma_{v, i}^2} = \frac{\sum_{n=1}^\infty n \cdot {\rm cov}\bigl( v_{k, i}, v_{k+n, i}\bigr) }{\sum_{n=1}^\infty {\rm cov}\bigl( v_{k,i}, v_{k+n,i} \bigr)} \ , 
 \label{eq:corr_time_def}
\end{align}
justifying the label correlation time for this variance ratio.
\end{theorem}
\smallskip

Here, $\sigma_{\theta, i}^2$ and $\sigma_{v, i}^2$ represent the variances of the weights and velocities, respectively, in a given direction $\vec{p}_i$, where $\theta_{k,i} \coloneqq \vec{\theta}_k \cdot \vec{p}_i$ and $v_{k,i} \coloneqq \vec{v}_k \cdot \vec{p}_i$. If the covariance matrices $\vec{\Sigma}$ and $\vec{\Sigma_v}$ commute, they can be simultaneously diagonalized, allowing us to choose shared eigenvectors $\vec{p}_i$. In this case, $\sigma_{\theta, i}^2$ and $\sigma_{v, i}^2$ become the corresponding eigenvalues. The detailed derivation is presented in \cref{sec:appendix_corr_time_calc}.

\subsection{Variance for late training phase}
\label{sec:var_main}

Given the autocorrelation of the noise calculated earlier, we aim to present the expected variances of the weights and velocities during the later stages of training. To characterize the conditions of this phase, we make the following assumptions.

{\bf Assumption 1: Quadratic approximation \hspace{1ex}}
We assume that we have reached a minimum point of the loss function, which  can be  adequately  represented with a quadratic form as \mbox{$L(\Vec{\theta}) = L_0 + \frac{1}{2}(\Vec{\theta} - \Vec{\theta_*})^\top\vec{H}(\Vec{\theta} - \Vec{\theta_*})$}. Without loss of generality, we  set $L_0=0 $ and $\Vec{\theta_*}=0$, simplifying the expression to $L(\Vec{\theta}) = \frac{1}{2}\Vec{\theta}^\top\vec{H}\Vec{\theta} \ .$

{\bf Assumption 2: Anti-correlated noise \hspace{1ex}}
We assume that the covariance of the SGD noise is static and that its autocorrelation follows the relation previously calculated in \cref{eq:noise_corr}, even when the weight vector is not static.

While the noise generally exhibits state dependence, it is reasonable to assume a constant covariance during the later stages of training and within the time frame of our analysis. As described by Ziyin et al.~\cite{Ziyin.2022}, state dependence enters the noise through the value of the training loss. Since we observe only minimal changes in the loss during our analysis period (see \cref{sec:appendix_loss_evolution}), assuming a static noise covariance is justified.

{\bf Assumption 3: Hessian and noise covariance commute \hspace{1ex}}
We assume that the covariance of the noise commutes with the Hessian matrix, $\vec{C}\vec{H}=\vec{H}\vec{C}$.

This assumption is not strictly necessary, but it simplifies the analysis. The theory predicts a variance reduction in the eigenspace of the Hessian with relatively small eigenvalues without requiring commutativity (see \cref{sec:appendix_commutation_assumption}). However, when $\vec{C}\vec{H}\neq\vec{H}\vec{C}$, the calculated weight and velocity variances are no longer exact eigenvalues of their respective covariance matrices; instead, they represent variances along the directions of the Hessian eigenvectors. As long as $\vec{C}$ and $\vec{H}$ approximately commute -- as discussed in \cref{sec:related_works_hessian_noise} -- these variances provide a good approximation of the actual eigenvalues.
For further discussion, we refer to \cref{sec:appendix_commutation_assumption}, where we demonstrate that although $\vec{C}\vec{H}=\vec{H}\vec{C}$ is not strictly satisfied in our empirical investigation, the eigenbasis of $\vec{H}$ remains a good approximation for the eigenbasis of $\vec{C}$. Moreover, due to finite sample sizes, the actual eigenvalues of the empirically recorded weight covariance matrix $\vec{\Sigma}$ are inherently skewed. Therefore, analyzing the weight variance in the eigenbasis of $\vec{H}$ is beneficial, as we explain in \cref{sec:appendix_feng_tu}.

Additionally, we assume that $0 \leq \beta < 1$ and $0 < \eta \lambda_i < 2(1 + \beta)$ for all eigenvalues $\lambda_i$ of $\vec{H}$. If these conditions are not met, the weight fluctuations would diverge. These bounds can be derived by requiring that the matrix which determines the dynamics in our model ($\vec{D_i}$ in \cref{eq:update_1d} and \cref{eq:master_matrix}) has all eigenvalues with magnitude strictly less than one.

{\bf Calculation for one eigenvalue \hspace{1ex}} With the previously stated assumptions in place, the covariance matrices $\vec{\Sigma}$ and $\vec{\Sigma_v}$ commute with $\vec{C}$, $\vec{H}$, and with each other (see \cref{sec:appendix_var_commutation}). As a result, they all share a common eigenbasis $\Vec{p_i}$, with $i=1, \dots, d$, which facilitates the  computation of the expected variance. We will outline the most important steps here, while details can be found in \cref{sec:appendix_var_exact}. For a given common eigenvector $\Vec{p_i}$, we project the relevant variables onto this vector. This yields the projected weight $\theta_{k,i} \coloneqq \Vec{p_i}\cdot\Vec{\theta}_k$, velocity $v_{k,i} \coloneqq \Vec{p_i}\cdot\Vec{v}_k$, and noise term $\delta g_{k,i}\coloneqq \Vec{p_i}\cdot\Vec{\delta g}_k$ at the update step $k$. Correspondingly, we define the eigenvalues for the common eigenvector $\vec{p_i}$ as $\lambda_i$ for $\Vec{H}$, $\sigma_{\theta, i}^2$ for $\Vec{\Sigma}$, $\sigma_{v, i}^2$ for $\Vec{\Sigma_v}$, and $\sigma_{\delta g, i}^2$ for $\Vec{C}$. We denote the number of batches per epoch as $M = N/S$, presuming it is an integer.

By introducing the vector $\vec{x}_{k, i} \coloneqq \bigl(\theta_{k, i} \quad \theta_{k-1, i} \bigr)^\top$ that contains not only the current weight variable but also the weight variable with a one-step time lag we can write the update equation as:
\begin{align}
\vec{x}_{k, i} &= \Vec{D_i}\Vec{x}_{k-1, i} - \eta \delta g_{k, i} \Vec{e}_1 \ , \label{eq:update_1d}
\end{align}
where $\vec{e}_1 \coloneqq \bigl( 1 \quad 0 \bigr)^\top$ and the matrix $\Vec{D_i}$ governs the deterministic part of the update. Together with its explicit expression, we further define a correlation term:
\begin{align}
 \Vec{D_i} &\coloneqq \begin{pmatrix} 1 + \beta - \eta \lambda_i & -\beta \\ 1 & 0 \end{pmatrix}  \ , \ \ \ \Vec{E_i} \coloneqq \Vec{D_i}\,{\rm cov}\left( \vec{x}_{k-1,i}, \delta g_{k,i}\Vec{e}_1 \right ) \ . \label{eq:master_matrix}
\end{align}
The term $\Vec{E_i}$ encapsulates the correlation between the current weight variable and the noise term of the next update step. Typically,  noise terms  are assumed to be  temporally uncorrelated, which would render $\Vec{E_i}$ null. However, given the anti-correlated nature of the noise, we find a non-zero  $\Vec{E_i}$. An explicit expression of $\Vec{E_i}$ can be found in \cref{sec:appendix_var_exact}.

\medskip
\begin{theorem}
\label{thm:var_exact}
With the above assumptions and definitions, the following relation for the weight and velocity variances holds:
\begin{align}
    \begin{pmatrix} \sigma_{\theta, i}^2 \\ \sigma_{v, i}^2 \end{pmatrix} &= \eta^2 \sigma_{\delta g, i}^2\Vec{F_i} \left[\Vec{e}_1  - \left(\vec{E_i} + \Vec{E_i}^\top\right)\Vec{e}_1 \right] \ ,
    \label{variances.eq}
\end{align}
where the matrix $\vec{F_i}$ is explicitly expressed as:
\begin{align}
 \vec{F_i} &= \frac{1}{(1-\beta)\bigl( 2(1+\beta) - \eta \lambda_i) \bigr)}\begin{pmatrix} \frac{1 + \beta}{\eta \lambda_i } & \frac{2\beta(\eta \lambda_i -1 - \beta)}{\eta \lambda_i} \\ 2 & 2(\eta \lambda_i - 2) \end{pmatrix} \, .
\end{align}
\end{theorem}
\smallskip

The calculations can be found in \cref{sec:appendix_var_exact}. The exact relation \cref{variances.eq} can be easily evaluated numerically but it can also be approximated by assuming that $ M \gg 1/ (1-\beta)$, which implies that the correlation time induced by momentum is substantially shorter than one epoch. Consequently, two distinct regimes of Hessian eigenvalues emerge, separated by $\lambda_{\textrm{cross}} \coloneqq 3 (1-\beta) / (\eta M)$. For each of these regimes, specific simplifications apply. As discussed below, the value of $\lambda_{\textrm{cross}}$ is derived as the point at which both approximations converge. As we will see, for eigenvalues below that value, the intrinsic correlation time induced by the loss landscape becomes longer than the timescale given by the anti-correlations.

%%%%%%%%%%%%%%%%%

\begin{figure}[ht]
    \centering
    \includegraphics[]{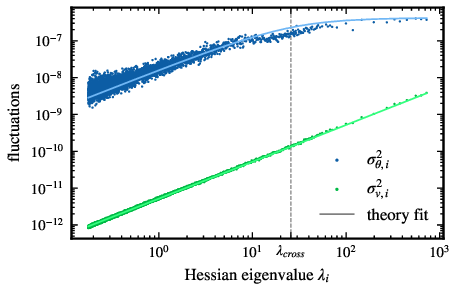}
    \includegraphics[]{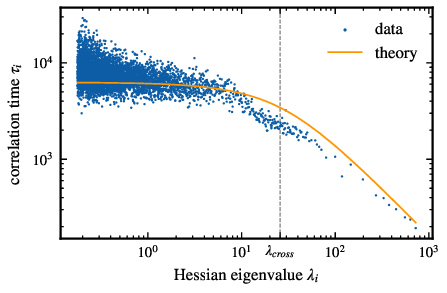}
    \caption{Relationship between Hessian eigenvalues and the variances of weights and velocities, as well as  correlation times. In the left panel, we present the variances of weights and velocities. The solid lines signify theoretical predictions from \cref{variances.eq}, assuming $\vec{C} \approx c_0 \vec{H}$ with $c_0$ fitted to the data. The right panel showcases the correlation time together  with  the  theoretical prediction resulting from \cref{variances.eq}, which does not require the $\vec{C} \approx c_0 \vec{H}$ assumption.}
    \label{fig:variance_and_tau}
\end{figure}

%%%%%%%%%%%%%%%%%%%%%%

\medskip
\begin{corollary}
\label{cor:large}
{\bf Relations for large Hessian eigenvalues:} For Hessian eigenvectors with eigenvalues $\lambda_i > \lambda_{\textrm{cross}}$ and when $M \gg 1/ (1-\beta)$,  the effects of noise anti-correlations are minimal. Consequently, we can use the following approximate relationships, which also hold true in the absence of correlations  and, therefore, also when drawing examples with replacement:
\begin{align}
    \sigma_{\theta, i}^2 \approx \frac{\eta^2\sigma_{\delta g, i}^2}{(1-\beta)\bigl(2(1+\beta) - \eta\lambda_i\bigr)} \cdot \frac{1+\beta}{\eta\lambda_i} \ , \
    \sigma_{v, i}^2 \approx \frac{\eta^2\sigma_{\delta g, i}^2}{(1-\beta)\bigl(2(1+\beta) - \eta\lambda_i\bigr)} \cdot 2 \ ,
\end{align}
and $\tau_i \approx \frac{1+\beta}{\eta\lambda_i}$ .
The detailed derivation of these formulas is presented in \cref{sec:appendix_var_approximation}.
\end{corollary}

It is frequently the case that the product $\eta \lambda_i$ is considerably less than one, enabling us to further simplify the prefactor of the variances. Assuming the noise covariance matrix is proportional to the Hessian matrix, such that $\sigma_{\delta g, i}^2 \propto \lambda_i$, we derive the following power laws for the variances: $\sigma_{\theta, i}^2 \propto \text{constant}$ and $\sigma_{v, i}^2 \propto \lambda_i$. As a result, in the subspace spanned by Hessian eigenvectors with eigenvalues $\lambda_i > \lambda_{\textrm{cross}}$, our theory predicts an isotropic weight variance $\vec{\Sigma}$.

\medskip
\begin{corollary}
\label{cor:small}
{\bf Relations for small Hessian eigenvalues:} In the case of Hessian eigenvectors associated with eigenvalues $\lambda_i < \lambda_{\textrm{cross}}$ and under the condition that $M \gg 1/ (1-\beta)$, the noise anti-correlation significantly modifies the outcome. We can express the approximate relationships as follows:
\begin{align}
    \sigma_{\theta, i}^2 \approx \frac{\eta^2\sigma_{\delta g, i}^2}{2(1-\beta)(1+\beta)} \cdot \frac{M}{3}\frac{1+\beta}{1-\beta} \ , \
    \sigma_{v, i}^2 \approx \frac{\eta^2\sigma_{\delta g, i}^2}{2(1-\beta)(1+\beta)} \cdot 2 
\end{align}
and $\tau_i \approx \frac{M}{3}\frac{1+\beta}{1-\beta} \eqqcolon \tau_{\textrm{SGD}}$ . Therefore, the weight variance is reduced by a factor of $\frac{M \eta \lambda_i}{3(1-\beta)}$ compared to the case without anti-correlations. The derivation of these formulas is provided in \cref{sec:appendix_var_approximation}.
\end{corollary}

If we once again assume that the noise covariance matrix is proportional to the Hessian matrix, such that $\sigma_{\delta g, i}^2 \propto \lambda_i$, we obtain the following power laws for the variances: $\sigma_{\theta, i}^2 \propto \lambda_i$ and $\sigma_{v, i}^2 \propto \lambda_i$. This implies that in the subspace spanned by Hessian eigenvectors with eigenvalues $\lambda_i < \lambda_{\textrm{cross}}$, the weight variance $\vec{\Sigma}$ is not isotropic but proportional to the Hessian matrix $\vec{H}$. It is noteworthy that the correlation time $\tau_i \approx \frac{M}{3}\frac{1+\beta}{1-\beta} \eqqcolon \tau_{\textrm{SGD}}$ in this subspace is independent of the Hessian eigenvalue. Moreover, $\tau_{\textrm{SGD}}$ is equivalent to the correlation time of the noise $M/3$, up to a factor that depends on momentum. 
Furthermore, with both approximations in place, one can understand the small-eigenvalue regime as the range of $\lambda$ for which the correlation time induced by the loss landscape exceeds the timescale of the anti-correlations -- both scaled appropriately by the momentum factor --i.e., when $\frac{1+\beta}{\eta\lambda} > \frac{M}{3} \frac{1+\beta}{1-\beta}$ for $\lambda < \lambda_{\rm cross}$.

\begin{table}[ht]
\centering
\begin{tabular}{|l|c|c|c|}
\hline
& weight variance $\sigma_{\theta,i}^2$ & velocity variance $\sigma_{v,i}^2$ & correlation time $\tau_i$ \\
\hline
\begin{tabular}[c]{@{}c@{}}large eigenvalues:\\ $\lambda_i > \lambda_{\rm cross}$\end{tabular} &
$\frac{1}{2} \cdot \frac{\eta^2\sigma_{\delta g, i}^2}{(1-\beta)(1+\beta)} \cdot \frac{1+\beta}{\eta\lambda_i}$ &
$\frac{\eta^2\sigma_{\delta g, i}^2}{(1-\beta)(1+\beta)}$ &
$\frac{1+\beta}{\eta\lambda_i}$ \\
\hline
\begin{tabular}[c]{@{}c@{}}small eigenvalues:\\ $\lambda_i < \lambda_{\rm cross}$\end{tabular} &
$\frac{1}{2} \cdot \frac{\eta^2\sigma_{\delta g, i}^2}{(1-\beta)(1+\beta)} \cdot \frac{M}{3} \frac{1+\beta}{1-\beta}$ &
$\frac{\eta^2\sigma_{\delta g, i}^2}{(1-\beta)(1+\beta)}$ &
$\frac{M}{3} \frac{1+\beta}{1-\beta} \coloneqq \tau_{\rm SGD}$ \\
\hline
\end{tabular}
\caption{
Summary of variances and correlation time along Hessian eigenvectors, depending on the eigenvalue magnitude. For simplicity, we assume $\vec{H}\vec{C} = \vec{C}\vec{H}$, $\eta\lambda_i \ll 1$ for all directions, and $M \gg 1/(1-\beta)$ with $M = N/S$ the number of minibatches per epoch. 
The results for large eigenvalues also coincide with those one obtains in the absence of anti-correlations in the noise. 
Note: fluctuations of the weight vector during an epoch can modify the noise anti-correlations and induce a minimal weight variance at very small eigenvalues, as discussed in \cref{sec:appendix_non_static_weights}.
}
\label{tab:variance_summary}
\end{table}

We summarize the results of Corollary~\ref{cor:large} and Corollary~\ref{cor:small} under simplifying assumptions in \cref{tab:variance_summary}. Notice that modifications to the noise anti-correlations caused by a fluctuating weight vector can affect the predictions above, particularly for very small Hessian eigenvalues. In \cref{sec:appendix_non_static_weights}, we show that these modifications introduce a minimal weight variance, below which the variance in very low-curvature directions cannot drop, even as the Hessian eigenvalues approach zero. However, in all our numerical experiments, the Hessian eigenvalues are sufficiently large that this lower-bound regime plays only a negligible role. In particular, in \cref{sec:appendix_non_static_weights} we derive that for the observed distribution of Hessian eigenvalues, the weight variance is expected to plateau only for eigenvalues $\lambda_i \lesssim 10^{-4}\lambda_{\rm max}$, which corresponds to values on the order of $10^{-1}$. In \cref{fig:variance_and_tau} we observe slight deviations from the predicted variance–curvature proportionality for eigenvalues on the order of $10^{-1}$.

\section{Numerics}
\subsection{Analysis Setup}
\label{sec:analysis_setup}

In order to corroborate our theoretical findings, we have conducted a small-scale experiment. 
We have trained a LeNet architecture, similar to the one described in \cite{Feng.2021}, using the CIFAR10 dataset \cite{Krizhevsky.2009}. LeNet is a compact convolutional network comprised of two convolutional layers followed by three dense layers. The network comprises approximately 137,000 parameters.
Here we present results for a single seed and specific hyperparameters. However, we have also performed tests with different seeds and combinations of hyperparameters, all of which showed comparable qualitative behavior (see \cref{sec:appendix_hyperparameters}). Furthermore, in \cref{sec:appendix_resnet20} we studied a ResNet architecture \cite{He.2016}, a more modern network, where we obtained similar results.

The training parameters together with the schedule are described in \cref{sec:appendix_loss_evolution} and result in a thousand minibatches per epoch,  $M = 1{,}000$. The setup achieves 95\% training accuracy and about 70\% testing accuracy. We compute the variances right after the initial schedule over a period of 20 additional epochs, equivalent to 20,000 update steps. 
During the initial schedule, we use an exponential learning rate decay applied once per epoch, while our theoretical results assume a constant learning rate. To better align the analysis with theory, we maintain a constant learning rate throughout the 20-epoch analysis period. However, since the learning rate changes only slowly across epochs -- and remains fixed within each epoch -- we expect deviations from theory due to this potential mismatch to be minimal when analyzing timeframes spanning only a few epochs.

The recorded weights are designated by $\Vec{\theta}_k$, with $k = K, \dots, K+T$ and $T = 20{,}000$. Given the impracticability of obtaining the full covariance matrix for all weights over this period due to the excessive memory requirements, we limit our analysis to a specific subspace. Employing the resource-efficient Pearlmutter trick \cite{Pearlmutter.1994}, we approximate the 5,000 largest eigenvalues and their associated eigenvectors of the Hessian matrix $\Vec{H}(\Vec{\theta}_K)$ at the beginning of the analysis period, drawn from the roughly 137,000 total. The eigenvectors of the Hessian matrix are represented by $\vec{p_i}$, and the projected weights  by $\theta_{k, i} = \Vec{\theta}_k \cdot \Vec{p_i}$. The variances are computed exclusively for these particular directions. The distribution of the approximated 5,000 eigenvalues is illustrated in \cref{sec:appendix_loss_evolution}.

\subsection{Noise Autocorrelations}
\label{sec:num_corr}
We scrutinize the correlations of noise by recording both the minibatch gradient $\Vec{g}_k(\Vec{\theta}_k)$ and the total gradient $\Vec{\nabla}L(\Vec{\theta}_k)$ at each update step throughout the analysis period, enabling us to capture the actual noise term $\vec{\delta g}_k(\Vec{\theta}_k)$. All these are projected onto the 5{,}000 approximated Hessian eigenvectors, yielding scalar components $\delta g_{k,i} = \vec{\delta g}_k(\Vec{\theta}_k) \cdot \vec{p}_i$. The theoretical prediction for the anti-correlation of the noise is a V-shaped structure proportional to the inverse of the number of batches per epoch, $1/M$, which, in our case, is on the order of $10^{-3}$. To extract the predicted relationship from the fluctuating data, we compute the autocorrelation for each direction $i$ individually as $\text{corr}_{\delta g, i}(h) = \frac{1}{\sigma_{\delta g, i}^2} \cdot \frac{1}{T - |h|} \sum_{k=K}^{T - |h|} \delta g_{k+|h|, i} \, \delta g_{k, i}$.

Due to the limited resolution from finite sampling, the standard deviation of each correlation estimate is around $1/\sqrt{T} \approx 7 \cdot 10^{-3}$ -- larger than the signal we aim to resolve. Therefore, we average the autocorrelation results across the 5,000  approximated eigenvectors, giving $\text{corr}_{\delta g,\, avg}(h)$ and reducing the expected noise level to approximately $1/\sqrt{5000 \cdot T} \approx 10^{-4}$, which enables detection of the predicted structure. \cref{fig:noise_corr} provides a visual representation of this analysis, showcasing a strong alignment between the empirical autocorrelation of noise and the prediction derived from our theory, confirming the V-shaped theoretical autocorrelation.

\subsection{Variances and Correlation time}
\label{sec:num_fluctuation}

Again, we limit our variance calculations to the directions of the 5,000 approximated Hessian eigenvectors. In the two different regimes of Hessian eigenvalues, either greater or lesser than the crossover value $\lambda_{\textrm{cross}}$, the weight and velocity variance closely follow the respective predictions from our theory (see left panel of \cref{fig:variance_and_tau}). 
We observe slight deviations from the predicted variance–curvature proportionality in the very low-curvature directions, where some weight variances are marginally higher than expected. These deviations may result from slow equilibration in very flat directions, as previous studies have observed that network weights continue to traverse the parameter space even after the loss appears to have stabilized \cite{Hoffer.2017, Feng.2021, Kunin.2021}. Another explanation could be corrections to the anti-correlations of the noise due to fluctuating weights. A more detailed discussion and extended analysis, can be found in \cref{sec:appendix_hessian_very_small}.
The calculated correlation time, derived from the ratio between the weight and velocity variance, also aligns well with our theoretical predictions (see right panel of \cref{fig:variance_and_tau}). This correlation time prediction remains independent of the exact relation between $\Vec{C}$ and $\Vec{H}$, thereby providing a more general result.

\section{Discussion}

We provide an intuitive interpretation of our results by examining the correlation times associated with different Hessian eigenvectors, which we also corroborate empirically. Although the 20-epoch analysis period may influence the observed correlation times -- since it is comparable to the maximum predicted correlation time $\tau_{\textrm{SGD}}$ -- we observe considerable differences when we sample examples without replacement during training (see \cref{sec:appendix_with_replacement}). This suggests that the observed correlation time behavior arises from the correlations introduced by sampling without replacement.

Throughout our analysis, we primarily focus on the weight covariance matrix $\mathbf{\Sigma}$, considering the correlation time $\tau_i$ and the velocity covariance $\mathbf{\Sigma_v}$ as auxiliary variables. Our results indicate that the velocity covariance $\mathbf{\Sigma_v}$ is directly proportional to the noise covariance $\mathbf{C}$. However, the behavior of the weight covariance $\mathbf{\Sigma}$ is more intricate due to the autocorrelation time, which connects the two variances. Specifically, for a single eigendirection, this relationship is given by $\sigma_{\theta, i}^2 = \frac{1}{2} \tau_i \sigma_{v, i}^2$. Examining these quantities in detail provides further insights. Moreover, our findings for these three quantities differ from those reported in a previous empirical study \cite{Feng.2021}. In \cref{sec:appendix_feng_tu}, we explain how the different analysis methods used in that study led to results impacted by finite-size effects.

In this manuscript, we examine the influence of smaller Hessian eigenvalues, which are often overlooked in discussions of minima characteristics in optimization landscapes. While prior work has focused on larger eigenvalues, we highlight the impact of reduced weight variance in flat directions. This observed reduction may help explain why SGD exploration predominantly occurs within a limited subspace of Hessian eigenvectors with higher curvature, as reported in previous studies \cite{Gur-Ari.2018, Xie.2021}.

We further connect our theoretical results to the long-standing hypothesis in deep learning that posits that flatter minima tend to generalize better than sharper ones. From this perspective, it is desirable for an optimization algorithm to escape sharp minima more efficiently than flat ones. To assess this behavior, we consider the implications of our result on the escaping efficiency $E_t \coloneqq \langle L(\vec{\theta}_t) - L(\vec{\theta}_0) \rangle_{\text{batch noise}}$, a concept introduced in prior work~\cite{Zhu.2019, Liu.2021}, which approximates the expected increase in loss due to stochastic weight fluctuations during training. Under a quadratic approximation of a minimum, $L(\vec{\theta}) \approx \frac{1}{2} (\vec{\theta} - \vec{\theta}_0)^{\rm T} \vec{H} (\vec{\theta} - \vec{\theta}_0) + L_0$, the stationary escaping efficiency becomes $E_\infty \approx \frac{1}{2} \mathrm{Tr}[\vec{H} \vec{\Sigma}]$, where $\vec{\Sigma}$ is the stationary weight covariance. For clarity, we consider our theoretical predictions for $\vec{C} = c_0 \vec{H}$ and without momentum, but similar qualitative behavior holds in the momentum case. When training with SGD with replacement, the weight variance is predicted to be isotropic, $\vec{\Sigma}^{\rm w.r.} = \frac{\eta c_0}{2} \vec{1}$, leading to an escaping efficiency $E_\infty^{\rm w.r.} = \frac{\eta c_0}{4} \mathrm{Tr}[\vec{H}]$. For SGD without replacement, however, the weight variance becomes anisotropic. In directions corresponding to eigenvalues $\lambda_i > \lambda_{\rm cross}$, the variance matches that of the with-replacement case, i.e., $\vec{\Sigma}_{\text{\tiny (\textgreater)}}^{\rm w/o. r.} = \frac{\eta c_0}{2} \vec{1}$. In contrast, for eigenvalues $\lambda_i < \lambda_{\rm cross}$, the variance becomes $\vec{\Sigma}_{\text{\tiny (\textless)}}^{\rm w/o. r.} = \frac{\eta c_0}{2} \frac{1}{\lambda_{\rm cross}} \vec{H}_{\text{\tiny (\textless)}}$, resulting in a reduced escaping efficiency $E_\infty^{\rm w/o. r.} = \frac{\eta c_0}{4} \mathrm{Tr}\big[\vec{H}_{\text{\tiny (\textgreater)}}\big] + \frac{\eta c_0}{4} \frac{1}{\lambda_{\rm cross}} \mathrm{Tr}\big[\vec{H}_{\text{\tiny (\textless)}}^2\big]$.
Since $\vec{H}_{\text{\tiny (\textless)}}$ represents the contribution from flatter directions (i.e., those with $\lambda_i < \lambda_{\rm cross}$), this result suggests that SGD without replacement is less prone to escaping flat minima than SGD with replacement -- particularly given that the majority of Hessian eigenvalues lie below the crossover value in our numerical experiments (see \cref{fig:hessian_density}). To demonstrate significance, we analyze the 5,000 largest Hessian eigenvalues of LeNet, include the effects of momentum, and find that our model predicts a 62\% reduction in escaping efficiency for the case without replacement compared to sampling with replacement. This effect would be even stronger if all 137,000 directions were considered. This selective suppression of variance in low-curvature directions may help the optimizer remain in flatter regions of the loss landscape, potentially contributing to improved generalization performance. Making this connection between suppressed variance and convergence towards flatter minima more explicit is a direction for future research.

To further assess the impact of anti-correlations in the gradient noise, we investigated the difference in generalization performance between drawing batches in SGD with and without replacement. We trained the network described in \cref{sec:analysis_setup} using the same training schedule and 20 different random seeds, considering the maximum test accuracy computed after each epoch. We found that training without replacement yielded test accuracies that were 0.8\% ± 0.2\% higher than training with replacement. Specifically, the maximum test accuracy for SGD without replacement was 69.8\% on average, compared to 69.0\% for SGD with replacement.

We relate this result directly to the anti-correlations we have described by comparing it with a prior study by Orvieto et al.~\cite{Orvieto.2022}. In that study, the authors considered full-batch gradient descent with artificially added noise that is anti-correlated in time. This noise was found to be beneficial for test accuracy and led to flatter minima. While the anti-correlations they considered have a very short correlation time, they are otherwise very similar to those inherent in SGD without replacement. We therefore propose that the positive effects described in their study could be extended to SGD without replacement due to these anti-correlations. Similarly, Beneventano~\cite{Beneventano.2023} studies SGD without replacement, showing that it regularizes the trace of the noise covariance along flat directions, leading to a faster escape from saddle points and a better generalization than SGD with replacement. Also, our findings on the impact of anti-correlated gradient noise in SGD without replacement relate to prior work on random reshuffling in convex optimization. In particular, G\"urb\"uzbalaban et al.~\cite{Gurbuzbalaban.2021} analyzes convergence rates under strong convexity and shows that without-replacement sampling can accelerate optimization.

\subsection*{Conclusion}

Our investigation of anti-correlations in SGD noise -- arising from drawing examples without replacement -- reveals a lower-than-expected weight variance in Hessian eigendirections with eigenvalues smaller than the crossover value $\lambda_{\textrm{cross}}$. By introducing the concepts of intrinsic correlation time, shaped by the loss landscape, and a constant noise correlation time, we provide deeper insights into the dynamics of SGD optimization. The reduced weight variance may allow gradients in flat directions to dominate fluctuations, steering the network toward even flatter minima. Training with anti-correlated noise can lead to improved generalization performance, suggesting that our findings may explain an important property of SGD that contributes to its success.

\section*{Note added}

After completion of this work, a study by Gross et al.~\cite{Gross.2024} derived a relation between weight variances and Hessian eigenvalues similar to the one discussed in this manuscript, making reference to the preprint \cite{Kuehn.2023} on which the present manuscript is based.

\newpage
\appendix
\FloatBarrier
\section{Definition of Limiting Quantities}
\label{sec:appendix_limit_definition}

When we speak of a covariance matrix or an average in the main text and in the following sections of the appendix, we mean the limiting average or the limiting covariance, unless otherwise specified. In other words, we are interested in the average of a quantity over one infinite run of SGD optimization, not the mean value for a fixed update step $k$ averaged over multiple runs of SGD optimization. With this in mind, we define the covariance matrix of two quantities $\vec{a}_k$ and $\vec{b}_k$ as
\begin{align}
    \textrm{cov}(\vec{a}_k, \vec{b}_k) \coloneqq \left\langle \left(\vec{a}_k - \langle \vec{a}_k \rangle_k\right) \left(\vec{b}_k - \langle \vec{b}_k \rangle_k\right)^\top\right\rangle_k
\end{align}
and the limiting average is defined as
\begin{align}
    \langle \vec{a}_k \rangle_k = \lim_{K \to \infty} \frac{1}{K+1}\sum_{k=k_0}^{k_0+K} \vec{a}_k \ .
\end{align}
When possible, we will suppress $k$ and denote the average as $\langle \cdot \rangle$. The average is independent of the starting value $k_0$, therefore we can shift indices within the average, meaning $\langle \vec{a}_k \rangle = \langle \vec{a}_{k+l} \rangle$ for any $l \in \mathbb{Z}$.

To see this we take any integer $l \in \mathbb{Z}$ and instead of adding it to the index $k$ we can also subtract it from the starting value $k_0$ and then separate the sum into two sums,
\begin{align}
    \langle \vec{a}_{k+l} \rangle &= \lim_{K \to \infty} \frac{1}{K+1}\sum_{k=k_0}^{k_0+K} \vec{a}_{k+l} \nonumber \\
    &= \lim_{K \to \infty} \frac{1}{K+1}\sum_{k=k_0 - l}^{k_0 - l +K} \vec{a}_{k} \nonumber \\
    &= \lim_{K \to \infty} \frac{1}{K+1}\sum_{k=k_0 - l}^{k_0 - 1} \vec{a}_{k}  + \lim_{K \to \infty}\frac{1}{K+1}\sum_{k=k_0}^{k_0 - l +K} \vec{a}_{k} \ .
\end{align}
The first sum is independent of $K$ except for the factor $\frac{1}{K+1}$, so the limit of the first part is zero. The second part of the limit can be rearranged as follows,
\begin{align}
    \lim_{K \to \infty}\frac{1}{K+1}\sum_{k=k_0}^{k_0 - l +K} \vec{a}_{k} &=  \lim_{K \to \infty}\frac{K-l+1}{K+1} \frac{1}{K-l+1}\sum_{k=k_0}^{k_0 - l +K} \vec{a}_{k} \nonumber \\
    &= \lim_{\tilde{K} \to \infty} \frac{1}{\tilde{K}+1}\sum_{k=k_0}^{k_0+\tilde{K}} \vec{a}_{k} \\
    &\eqqcolon \langle \vec{a}_k \rangle \ ,
\end{align}
where we used $\lim_{K \to \infty}\frac{K-l+1}{K+1} = 1$ and renamed $K - l$ in the second to last step. All together this gives us the desired relation $\langle \vec{a}_k \rangle = \langle \vec{a}_{k+l} \rangle$.

If one were to consider the covariance for a fixed update step $k$ averaged over multiple runs of SGD optimization, it is possible that this covariance could depend on the index $k$, but this is not our case of interest.

%%%%%%%%%%%%%%%%%%%%%%%%%%%%%%%%%%%%%%%%%%%%%%%%%%
%%%%%%%%%%%%%%%%%%%%%%%%%%%%%%%%%%%%%%%%%%%%%%%%%%
%%%%%%%%%%%%%%%%%%%%%%%%%%%%%%%%%%%%%%%%%%%%%%%%%%
\FloatBarrier
\section{Variance Calculation}
\label{sec:appendix_var_calc}

\subsection{Commutativity of the covariance matrices}
\label{sec:appendix_var_commutation}

In this section, we show that if $\vec{C}\vec{H}=\vec{H}\vec{C}$ also $\vec{\Sigma}$ and $\vec{\Sigma_v}$ will commute with $\vec{C}$, with $\vec{H}$ and with each other. We make the assumptions one to three from \cref{sec:var_main} and therefore the SGD update equations become
\begin{align}
    \Vec{v}_{k} &=  -\eta \Vec{H} \Vec{\theta}_{k-1} + \beta \Vec{v}_{k-1} - \eta\Vec{\delta g}_k \ ,\\
    \vec{\theta}_{k} &= \left(\Vec{1} -\eta \Vec{H} \right) \Vec{\theta}_{k-1} + \beta \Vec{v}_{k-1} - \eta\Vec{\delta g}_k \ ,
\end{align}
which can be rewritten by using the vector $\Vec{y}_k \coloneqq \bigl( \vec{\theta}_{k} \quad \Vec{v}_{k} \bigr)^\top$, combining both the current weight and velocity variable, to be
\begin{align}
    \Vec{y}_{k+1} = \Vec{X} \Vec{y}_{k} - \Vec{z}_{k+1}   \ . \label{eq:var_proof_commute_master}
\end{align}
Here,  $\Vec{z}_k \coloneqq \bigl( \eta\Vec{\delta g}_k \quad \eta\Vec{\delta g}_k \bigr)^\top$ contains the current noise term, and the matrix governing the deterministic part of the update is defined to be 
\begin{align}
    \Vec{X} \coloneqq \begin{pmatrix} \Vec{1} -\eta \Vec{H} & \beta \Vec{1} \\ -\eta \Vec{H} & \beta \Vec{1} \end{pmatrix} \ .
\end{align}

By iteratively applying \cref{eq:var_proof_commute_master} we obtain
\begin{align}
    \Vec{y}_{k+h} = \Vec{X}^h \Vec{y}_{k} - \sum_{i=1}^{h} \Vec{X}^{h-i} \Vec{z}_{k+i} \ . \label{eq:var_proof_commute_master_iter}
\end{align}
Under the assumption  $0 \leq \beta < 1$ and $0 < \eta \lambda_i < 2(1 + \beta)$, for all eigenvalues $\lambda_i$ of $\Vec{H}$, the magnitude of the eigenvalues of $\vec{X}$ will be less than one. It is straightforward to show this relation for the eigenvalues of $\Vec{X}$ by using the eigenbasis of $\Vec{H}$. Therefore,
\begin{align}
    \lim_{h \rightarrow \infty} \Vec{X}^h \Vec{y}_{k} = 0. \label{eq:var_proof_commute_zero_mean}
\end{align}
As we can shift the index in the weight variance, $h$ can be chosen arbitrarily large, which yields the following relation for the covariance
\begin{align}
    \langle \Vec{y}_k \Vec{y}_k^\top \rangle = \lim_{h \rightarrow \infty} \sum_{i, j=1}^{h} \Vec{X}^{h-i} \langle \Vec{z}_{k+i} \Vec{z}_{k+j}^\top \rangle \left( \Vec{X}^{h-j} \right)^\top.
    \label{eq:var_proof_commute_limit}
\end{align}

Because \cref{eq:var_proof_commute_master_iter} together with \cref{eq:var_proof_commute_zero_mean} implies $\langle \vec{y}_{k} \rangle = 0$ and therefore  $\langle \Vec{\theta}_{k} \rangle = 0$ and $\langle \Vec{v}_{k} \rangle = 0$, the left hand side of \cref{eq:var_proof_commute_limit} contains the covariance matrices of interest,
\begin{align}
     \langle \Vec{y}_k \Vec{y}_k^\top \rangle = \begin{pmatrix} \Vec{\Sigma} & \langle\Vec{\theta}_k \Vec{v}_k^\top \rangle \\ \langle\Vec{v}_k \Vec{\theta}_k^\top \rangle & \Vec{\Sigma_v} \end{pmatrix} \ .  \label{eq:var_proof_commute_covariance}
\end{align}
From \cref{eq:var_proof_commute_limit} we can also infer that $\langle \Vec{y}_k \Vec{y}_k^\top \rangle$ is finite as the magnitude of the eigenvalues of $\vec{X}$ is less than one. Consequently, by \cref{eq:var_proof_commute_covariance}, the covariance matrices $\vec{\Sigma}$ and $\Vec{\Sigma_v}$ are finite as well. The average over the noise terms $\Vec{z}_k$ on the right hand side of \cref{eq:var_proof_commute_limit} is by assumption equal to
\begin{align}
     \langle \Vec{z}_{k+i} \Vec{z}_{k+j}^\top \rangle = \eta^2\left( \delta_{i,j} -\mathbf{1}_{\{1, ..., M\}}(|i-j|) \frac{M-|i-j|}{M(M-1)} \right) \cdot \begin{pmatrix} \Vec{C} & \Vec{C} \\ \Vec{C} & \Vec{C} \end{pmatrix} \ ,
\end{align}
from which it follows that for any finite $h$ every matrix entry of the two by two super matrix on the right hand side of \cref{eq:var_proof_commute_limit} is a function of $\vec{C}$ and $\vec{H}$. Therefore, when considering the limit $h \rightarrow \infty$, $\vec{C}\vec{H}=\vec{H}\vec{C}$ implies that $\Vec{\Sigma}$ and $\Vec{\Sigma_v}$ will also commute with $\vec{C}$, with $\vec{H}$ and with each other.

\subsection{Proof of the variance  formula for one specific eigenvalue}
\label{sec:appendix_var_exact}

Since $\vec{\Sigma}$ and $\vec{\Sigma_v}$ will commute with $\vec{C}$, with $\vec{H}$ and with each other, it is sufficient to prove the one dimensional case. For the multidimensional case simply apply the proof in the direction of each common eigenvector individually. 
The expectation values discussed below are computed with respect to the asymptotic distributions of $\theta$ and $v$,  since  we are only interested in the asymptotic behavior of training. We want to find $\sigma_\theta^2 \coloneqq \langle \theta_k \theta_k \rangle$ and $\sigma_v^2 \coloneqq \langle v_k v_k \rangle$. We assume $0 \leq \beta < 1$ and $0 < \eta \lambda < 2(1 + \beta)$ where $\lambda$ is the Hessian eigenvalue.

The equations describing SGD in one dimension are: 
\begin{align}
    g_k(\theta) &= \frac{\partial}{\partial \theta} L(\theta) + \delta g_k(\theta)\\
    v_{k} &=  -\eta g_{k}(\theta_{k-1}) + \beta v_{k-1}\\
    \theta_{k} &= \theta_{k-1} + v_{k} \ .
\end{align}

Our remaining assumptions can then be described the following way
\begin{align}
    L(\theta) &= \frac{1}{2}\theta \lambda \theta\\
    \delta g_k(\theta) &=  \delta g_k \\ 
    \langle \delta g_k \delta g_{k+h} \rangle &= \sigma_{\delta g}^2 \left( \delta_{h,0} -\mathbf{1}_{\{1, ..., M\}}(|h|) \frac{M-|h|}{M(M-1)} \right) \label{eq:var_proof_noise_assumption}\\
    \sigma_{\delta g}^2 &\coloneqq  \langle \delta g_k \delta g_{k} \rangle \ .
\end{align}

With these assumptions the update equations can be described by a discrete stochastic linear equation of second order
\begin{align}
    \theta_k = (1+\beta - \eta \lambda ) \theta_{k-1} - \beta \theta_{k-2} - \eta \delta g_k
\end{align}
which can be rewritten into matrix form as follows
\begin{align}
    \vec{x}_k &= \Vec{D}\Vec{x}_{k-1} - \eta \delta g_k \Vec{e}_1 \label{eq:var_proof_master}\\
    \vec{x}_k &\coloneqq \begin{pmatrix} \theta_k \\ \theta_{k-1} \end{pmatrix}\\
    \vec{e}_1 &\coloneqq \begin{pmatrix} 1 \\ 0 \end{pmatrix}\\
    \Vec{D} &\coloneqq \begin{pmatrix} 1 + \beta - \eta \lambda & -\beta \\ 1 & 0 \end{pmatrix} \ . \label{eq:appendix_master_matrix}
\end{align}

We are now interested in the following covariance matrix
\begin{align}
    \Vec{\tilde{\Sigma}} &\coloneqq \left\langle \Vec{x}_k \Vec{x}_k^\top \right\rangle \label{eq:var_proof_covariance} \nonumber\\
    &= \begin{pmatrix} \sigma_\theta^2 & \langle \theta_k \theta_{k-1} \rangle \\ \langle\theta_k \theta_{k-1} \rangle  & \sigma_\theta^2 \end{pmatrix}
\end{align}
where the second equality is due to the fact that $\langle \theta_k \theta_k \rangle = \langle \theta_{k-1} \theta_{k-1} \rangle$. 
{ As we are interested in the asymptotic covariance, this expectation value is independent of any finite shift of the index $k$.} By inserting \cref{eq:var_proof_master} into $\left\langle \Vec{x}_k \Vec{x}_k^\top \right\rangle$ we arrive at the following equality
\begin{align}
    \left\langle \Vec{x}_k \Vec{x}_k^\top \right\rangle = \Vec{D}\left\langle \Vec{x}_{k-1} \Vec{x}_{k-1}^\top \right\rangle \Vec{D}^\top + \eta^2\left\langle\delta g_k \delta g_k \right\rangle \Vec{e}_1 \Vec{e}_1^\top - \eta \left(\Vec{D}\left\langle \vec{x}_{k-1} \delta g_k \right \rangle \Vec{e}_1^\top + \left(\Vec{D}\left\langle \vec{x}_{k-1} \delta g_k \right \rangle \Vec{e}_1^\top\right)^\top \right)
\end{align}
which can be simplified to the equivalent equation
\begin{align}
    \Vec{\tilde{\Sigma}} - \Vec{D}\Vec{\tilde{\Sigma}}\Vec{D}^\top = \eta^2\sigma_{\delta g}^2 \Vec{e}_1 \Vec{e}_1^\top - \eta \left(\Vec{D}\left\langle \vec{x}_{k-1} \delta g_k \right \rangle \Vec{e}_1^\top + \left(\Vec{D}\left\langle \vec{x}_{k-1} \delta g_k \right \rangle \Vec{e}_1^\top\right)^\top \right) \ . \label{eq:var_proof_sigma_simplified}
\end{align}
If we apply the left-hand side on the vector $\Vec{e}_1$, it can be expressed as
\begin{align}
    \left[ \Vec{\tilde{\Sigma}} - \Vec{D}\Vec{\tilde{\Sigma}}\Vec{D}^\top \right] \Vec{e}_1 &= \Vec{F}_1^{-1}\Vec{\tilde{\Sigma}}\Vec{e}_1 \label{eq:var_proof_sigma_inverted}\\
    \Vec{F}_1^{-1} &\coloneqq \begin{pmatrix} \eta \lambda (2-\eta \lambda) -2\beta(1+\beta - \eta \lambda) & 2\beta(1+\beta - \eta \lambda) \\ -(1+\beta - \eta \lambda) & 1+\beta \end{pmatrix} \ .
\end{align}
Also notice $v_k = \theta_k - \theta_{k-1}$ and therefore
\begin{align}
    \sigma_v^2 = 2\sigma_\theta^2 - 2\langle \theta_k \theta_{k-1} \rangle \ ,
\end{align}
again due to the fact that the expectation value does not depend on k. Hence, the variances can then be expressed as
\begin{align}
    \begin{pmatrix} \sigma_\theta^2 \\ \sigma_v^2 \end{pmatrix}
    & = \Vec{F}_2 \Vec{\tilde{\Sigma}} \Vec{e}_1 \label{eq:var_proof_double_var}\\
    \Vec{F}_2 &\coloneqq \begin{pmatrix} 1 & 0 \\ 2 & -2\end{pmatrix} \ .
\end{align}
We define the matrix $\Vec{F} \coloneqq \Vec{F}_2 \Vec{F}_1$. By applying both sides of \cref{eq:var_proof_sigma_simplified} to the vector $\Vec{e_1}$, then multiplying by the matrix $\Vec{F}$ from the left and using \cref{eq:var_proof_sigma_inverted,eq:var_proof_double_var} we obtain
\begin{align}
    \begin{pmatrix} \sigma_\theta^2 \\ \sigma_v^2 \end{pmatrix} = \Vec{F} \left[ \eta^2\sigma_{\delta g}^2 \Vec{e}_1 \Vec{e}_1^\top - \eta \left(\Vec{D}\left\langle \vec{x}_{k-1} \delta g_k \right \rangle \Vec{e}_1^\top + \left(\Vec{D}\left\langle \vec{x}_{k-1} \delta g_k \right \rangle \Vec{e}_1^\top\right)^\top \right) \right]\Vec{e}_1  \ . \label{eq:var_proof_var_simplified}
\end{align}
with
\begin{align}
    \vec{F} = \frac{1}{(1-\beta)\bigl( 2(1+\beta) - \eta \lambda) \bigr)}\begin{pmatrix} \frac{1 + \beta}{\eta \lambda } & \frac{2\beta(\eta \lambda -1 - \beta)}{\eta \lambda} \\ 2 & 2(\eta \lambda - 2) \end{pmatrix} \ .
\end{align}

To simplify \cref{eq:var_proof_var_simplified} further we go back to \cref{eq:var_proof_master} and iterate it to obtain
\begin{align}
    \vec{x}_k = \Vec{D}^n\vec{x}_{k-n} - \eta \sum_{h=0}^{n-1} \Vec{D}^h\Vec{e}_1 \delta g_{k-h} \ .
\end{align}
We note that $\left\langle \vec{x}_{k-n} \delta g_k \right \rangle = 0$ for $ n \geq M$. The correlation between noise terms separated by at least one epoch vanishes, and $\vec{x}_{k}$ only depends on past noise terms. By setting $n = M$ we find
\begin{align}
    \left\langle \vec{x}_{k-1} \delta g_k \right \rangle &= \Vec{D}^M\left\langle \vec{x}_{k-1-M} \delta g_k \right \rangle - \eta \sum_{h=0}^{M-1} \Vec{D}^h\Vec{e}_1 \left\langle \delta g_k \delta g_{k-1-h}\right \rangle \nonumber\\
    &= -\eta \sigma_{\delta g}^2 \sum_{h=0}^{M-1} \Vec{D}^h\Vec{e}_1 \left( -\frac{M-(h+1)}{M(M-1)} \right) \ ,\label{eq:var_proof_geometric_series}
\end{align}
where the assumption about the correlation of the noise terms, \cref{eq:var_proof_noise_assumption}, was inserted for the last line. \cref{eq:var_proof_geometric_series} is a sum of a finite geometric series and a derivative of that which can be simplified to
\begin{align}
    \left\langle \vec{x}_{k-1} \delta g_k \right \rangle &= \eta \sigma_{\delta g}^2 \frac{\vec{D}^M + (\Vec{1}-\Vec{D})M - \Vec{1}}{(\Vec{1}-\Vec{D})^2M(M-1)} \,\Vec{e}_1 \ .
\end{align}
Substituting this result back into \cref{eq:var_proof_var_simplified} yields
\begin{align}
    \begin{pmatrix} \sigma_\theta^2 \\ \sigma_v^2 \end{pmatrix} &= \eta^2 \sigma_{\delta g}^2\Vec{F} \left[\Vec{e}_1  - \left(\vec{E} + \Vec{E}^\top\right)\Vec{e}_1 \right] \label{eq:var_proof_var_exact}
\end{align}
with the definition
\begin{align}
    \Vec{E} &\coloneqq \Vec{D}\frac{\vec{D}^M + (\Vec{1}-\Vec{D})M - \Vec{1}}{(\Vec{1}-\Vec{D})^2M(M-1)} \,\Vec{e}_1\Vec{e}_1^\top \ .
\end{align}
With \cref{eq:var_proof_var_exact} we have arrived at the exact formula for the variances which can easily be evaluated numerically.

\subsection{Approximation of the exact formula}
\label{sec:appendix_var_approximation}

It is possible to approximate the exact result for the variance assuming small or large eigenvalues, respectively. For that, it is necessary to approximate $\vec{D}^M\vec{e_1}$. To do so, we will use the following eigendecomposition of $\vec{D}$
\begin{align}
    \Vec{D} &= \Vec{Q}\Vec{\Lambda}\Vec{Q}^{-1}\\
    \Vec{\Lambda} &= \begin{pmatrix} \Lambda_+ & 0\\ 0 & \Lambda_- \end{pmatrix}\\
    \Vec{Q} &= \begin{pmatrix} \Lambda_+ & \Lambda_-\\ 1 & 1 \end{pmatrix}\\
    \Vec{Q}^{-1} &= \frac{1}{\Lambda_+ - \Lambda_-} \begin{pmatrix} 1 & -\Lambda_-\\ -1 & \Lambda_+ \end{pmatrix}\\
    \Lambda_{\pm} &= \frac{1}{2}\left(1+\beta - \eta \lambda \pm s \right)\\
    s &\coloneqq \sqrt{\left(1-\beta\right)^2 - \eta \lambda \bigl(2(1+\beta) -\eta \lambda\bigr)}
\end{align}
It is straightforward to show that the magnitude of the eigenvalues of $\Vec{D}$ is strictly smaller than one, $|\Lambda_\pm| < 1$, under the conditions $0 < \eta \lambda < 2(1 +\beta)$ and $0 \leq \beta < 1$.

\subsubsection*{Large Hessian eigenvalues}
\begin{align}
    \sigma_{\theta}^2 &\approx \frac{\eta^2\sigma_{\delta g}^2}{(1-\beta)(2(1+\beta) - \eta \lambda)} \cdot \frac{1 +\beta}{\eta \lambda} \\
    \sigma_{v}^2 &\approx \frac{\eta^2\sigma_{\delta g}^2}{(1-\beta)(2(1+\beta) - \eta \lambda)} \cdot 2
\end{align}

We will show that this approximation for large Hessian eigenvalues is valid under the assumption $ M (\eta \lambda)^2 \gg 1$ where $M$ is the number of batches per epoch. However, numerical studies indicate that these relations also hold under the previously mentioned assumptions of $\frac{M \eta \lambda }{1 - \beta } \gg 1$, equivalent to $\lambda \gtrsim \lambda_\textrm{cross}$, and $M (1-\beta) \gg 1$.

Inserting the eigendecomposition of $\vec{D}$ into the expression $\Vec{D}^M\Vec{e_1}$ yields
\begin{align}
    \Vec{D}^M\Vec{e_1} = \begin{pmatrix} y_{M+1} \\ y_{M} \end{pmatrix}\\
    y_M \coloneqq \frac{\Lambda_+^M - \Lambda_-^M}{\Lambda_+ - \Lambda_-}.
\end{align}
From the definition of $y_M$ one sees that
\begin{align}
    y_M = \frac{\Lambda_+ + \Lambda_-}{2}\,y_{M-1} + \frac{\Lambda_+^{M-1} + \Lambda_-^{M-1}}{2}  \ ,
\end{align}
and by using $|\Lambda_\pm| < 1$ as well as $y_0 = 0$ one can show iteratively that
\begin{align}
    |y_M| \leq M+1.
\end{align}
Therefore, we have
\begin{align}
    \left\lVert \Vec{D}^M\Vec{e_1} \right\rVert_\infty \leq M + 1
\end{align}
where $\left\lVert \cdot \right\rVert_\infty$ is denoting the maximum norm $\left\lVert \vec{x} \right\rVert_\infty \coloneqq \max_i |x_i|$ for a vector $\Vec{x}$ or its induced matrix norm $\left\lVert \vec{A} \right\rVert_\infty \coloneqq \max_i \sum_j |a_{ij}|$ for a matrix $\Vec{A}$.

Explicit calculations show that 
\begin{align}
    \left\lVert (\vec{1} - \Vec{D})^{-1} \right\rVert_\infty \leq \frac{4}{\eta \lambda}
\end{align} under the assumption that $0 \leq \beta < 1$ and $0 < \eta \lambda < 2(1 + \beta)$. From here it is straightforward to show that 
\begin{align}
    \left\lVert \left(\Vec{E} + \Vec{E}^\top\right)\Vec{e_1} \right\rVert_\infty \leq \frac{\tilde{c}}{M (\eta \lambda)^2}
\end{align}
where $\Tilde{c}$ is a factor of order unity under the constraints $0 \leq \beta < 1$ and $0 < \eta \lambda < 2(1 + \beta)$. By substituting this result back into \cref{eq:var_proof_var_exact} one directly  sees that a comparison to the approximation yields
\begin{align}
    \left\lvert 1 - \frac{\sigma_\theta^2}{\sigma_{\theta, \textrm{large}}^2} \right\rvert \leq \frac{c_1}{M (\eta \lambda)^2} \ ,\\
    \left\lvert 1 - \frac{\sigma_v^2}{\sigma_{v, \textrm{large}}^2} \right\rvert \leq \frac{c_2}{M (\eta \lambda)^2} \ ,
\end{align}
where $c_1$ and $c_2$ are again of order unity and the approximation is defined as
\begin{align}
    \begin{pmatrix} \sigma_{\theta, \textrm{large}}^2 \\ \sigma_{v, \textrm{large}}^2 \end{pmatrix} &\coloneqq \eta^2 \sigma_{\delta g}^2\Vec{F} \Vec{e}_1 \nonumber\\ &=\frac{\eta^2\sigma_{\delta g}^2}{(1-\beta)(2(1+\beta) - \eta \lambda)} \cdot \begin{pmatrix} \frac{1 +\beta}{\eta \lambda}  \\ 2 \end{pmatrix}.
\end{align}
Interestingly, one can see that the approximation for large Hessian eigenvalues is equivalent to the result we would obtain if we assumed there was no autocorrelation of the noise to begin with.

In the case where  the stricter assumption is not true, $M(\eta\lambda)^2<1$, but the numerically obtained conditions still hold, $\lambda \gtrsim \lambda_{\textrm{cross}}$ and $M(1-\beta) \gg 1$, it occurs that $\left\lVert \left(\Vec{E} + \Vec{E}^\top\right)\Vec{e_1} \right\rVert_\infty$ is no longer small. But in that case, $\Vec{F}\left(\Vec{E} + \Vec{E}^\top\right)\Vec{e_1}$ can still be neglected compared to $\Vec{F}\Vec{e_1}$, as numerical experiments show. 

\subsubsection*{Small Hessian eigenvalues}

To obtain the relations for small Hessian eigenvalues, we perform a Taylor expansion with respect to $\lambda$ with the help of computer algebra. We neglect the terms which are at least of order $\lambda$. Numerical study indicates that these relations hold under the mentioned assumption of $\lambda \lesssim \lambda_\textrm{cross}$ and \mbox{$M (1-\beta) \gg 1$}.

It is straightforward but lengthy to obtain the following expression using the eigendecomposition of $\Vec{D}$
\begin{align}
    \begin{pmatrix} \sigma_\theta^2 \\ \sigma_v^2 \end{pmatrix} = \frac{\eta^2\sigma_{\delta g}^2}{2(1-\beta)(1+\beta)} \cdot \begin{pmatrix} \frac{M}{3}\frac{1+\beta}{1-\beta} + \mathcal{O}(\lambda) \\ 2 + \mathcal{O}(\lambda) \end{pmatrix}
\end{align}
where the zeroth order terms are simplified under approximation $M (1-\beta) \gg 1$.

\subsection{Satisfying the assumptions of the correlation time relation}
\label{sec:appendix_var_tau_assumptions}

In this section we want to show that the weight and velocity variances resulting from stochastic gradient descent as described above and in \cref{sec:var_main} satisfies the necessary assumptions (i) to (iii) of \cref{thm:correlation_time} such that the velocity correlation time is equal to $\tau_i = 2\sigma_{\theta, i}^2 / \sigma_{v, i}^2$. Validity of assumption (i) existence and finiteness of $\vec{\Sigma} \coloneqq $ cov$(\vec{\theta}_k, \vec{\theta}_k)$, $\vec{\Sigma_v} \coloneqq $ cov$(\vec{v}_k, \vec{v}_k)$, and $\langle \vec{\theta} \rangle$ can be inferred from the calculation presented in \cref{sec:appendix_var_commutation}. Therefore, we concentrate on assumption (ii) $\lim_{n \to \infty}\textrm{cov}\bigr(\vec{\theta}_{k}, \vec{\theta}_{k+n}\bigr) = 0$ and (iii) $\lim_{n \to \infty}n\cdot\textrm{cov}\bigr(\vec{\theta}_{k}, \vec{\theta}_{k+n} - \vec{\theta}_{k+n+1}\bigr) = 0$. We consider the one dimensional case, but the extension to the multidimensional case is straightforward. Additionally $\langle \Vec{\theta}\rangle = 0$ (see \cref{sec:appendix_var_commutation}) and, therefore, the remaining two assumptions (ii) and (iii) can be written as  $\lim_{m \rightarrow \infty}{\langle\theta_k \theta_{k+m} \rangle} = 0$ and $\lim_{m \rightarrow \infty}{m\bigl(\langle\theta_k \theta_{k+m} \rangle - \langle\theta_k \theta_{k+m+1} \rangle\bigr)} = 0$. 

We will now show that for stochastic gradient descent under the assumptions of \cref{sec:var_main} the more restrictive relation $\lim_{m \rightarrow \infty}{m\langle\theta_k \theta_{k+m} \rangle} = 0$ is satisfied, from which follows (ii) and (iii). Following \cref{sec:appendix_var_exact} and using the same notation, we have the relation
\begin{align}
    \langle \Vec{x}_k \Vec{x}_{k-m}^\top \rangle = \vec{D}  \langle \Vec{x}_{k-1} \Vec{x}_{k-m}^\top \rangle - \eta \Vec{e_1} \langle  \delta g_k \Vec{x}_{k-m}^\top \rangle  \ , \label{eq:appendix_var_tau_assumption_1}
\end{align}
where $\Vec{x}_k \coloneqq \bigl( \theta_k \quad \theta_{k-1} \bigr)^\top$. For $m>M$, with $M$ being the number of batches per epoch, the correlation with the noise term on the right hand side of \cref{eq:appendix_var_tau_assumption_1} is equal to zero as discussed in \cref{sec:appendix_var_exact}. By iterating \cref{eq:appendix_var_tau_assumption_1}, for $m>M$ we have 
\begin{align}
    \langle \Vec{x}_k \Vec{x}_{k-m}^\top \rangle &= \vec{D}^{m-M-1}  \langle \Vec{x}_{k-m+M+1} \Vec{x}_{k-m}^\top \rangle \nonumber\\
    &= \vec{D}^{m-M-1}  \langle \Vec{x}_{k} \Vec{x}_{k-M-1}^\top \rangle \ .
    \label{eq:appendix_var_tau_assumption_2}
\end{align}
As described in \cref{sec:appendix_var_exact}, the magnitude of both eigenvalues of $\Vec{D}$ is strictly smaller than one. This implies that there exists a matrix norm $\left\lVert \cdot \right\rVert_{\Vec{D}}$ such that $\left\lVert \Vec{D} \right\rVert_{\Vec{D}} < 1$ from which one can deduce
\begin{align}
    \left\lVert \langle \Vec{x}_k \Vec{x}_{k-m}^\top \rangle \right\rVert_{\Vec{D}} \leq \left\lVert \vec{D} \right \rVert_{\Vec{D}}^{m-M-1} \cdot \left\lVert \langle \Vec{x}_{k} \Vec{x}_{k-M-1}^\top \rangle \right \rVert_{\Vec{D}} \ .
    \label{eq:appendix_var_tau_assumption_3}
\end{align}
Taking the limit of $m \to \infty$ we obtain
\begin{align}
    \lim_{m \to \infty} m \left\lVert \langle \Vec{x}_k \Vec{x}_{k-m}^\top \rangle \right\rVert_{\Vec{D}} &\leq \textrm{const} \cdot \lim_{m \to \infty} m \left\lVert \vec{D} \right \rVert_{\Vec{D}}^{m-M-1} \nonumber\\
    &= 0  \ ,
    \label{eq:appendix_var_tau_assumption_4}
\end{align}
and because
\begin{align}
    \langle \Vec{x}_k \Vec{x}_{k-m}^\top \rangle = \begin{pmatrix} \langle \theta_k \theta_{k-m} \rangle & \langle \theta_k \theta_{k-m-1} \rangle \\ \langle \theta_{k-1} \theta_{k-m} \rangle & \langle \theta_{k-1} \theta_{k-m-1} \rangle\end{pmatrix}
    \label{eq:appendix_var_tau_assumption_5}
\end{align}
we finally find
\begin{align}
    \lim_{m \rightarrow \infty}{m\langle\theta_k \theta_{k-m} \rangle} &= 0 \nonumber\\
    \Rightarrow \lim_{m \rightarrow \infty}{m\langle\theta_k \theta_{k+m} \rangle} &= 0 \ .
    \label{eq:appendix_var_tau_assumption_6}
\end{align}

%%%%%%%%%%%%%%%%%%%%%%%%%%%%%%%%%%%%%%%%%%%%%%%%%%
%%%%%%%%%%%%%%%%%%%%%%%%%%%%%%%%%%%%%%%%%%%%%%%%%%
%%%%%%%%%%%%%%%%%%%%%%%%%%%%%%%%%%%%%%%%%%%%%%%%%%
\section{Calculation of the Correlation Time Relation}
\label{sec:appendix_corr_time_calc}

We want to prove \cref{thm:correlation_time}, that is, the relation 
\begin{align}
    \frac{2 \sigma_{\theta, i}^2}{\sigma_{v, i}^2} = \frac{\sum_{n=1}^\infty n \, \langle v_{k, i} v_{k+n, i} \rangle}{\sum_{n=1}^\infty \langle v_{k,i} v_{k+n,i} \rangle} \ ,
\end{align}

under the following three assumptions: (i) Existence and finiteness of \mbox{$\vec{\Sigma} \coloneqq $ cov$(\vec{\theta}_k, \vec{\theta}_k)$,} \mbox{$\vec{\Sigma_v} \coloneqq $ cov$(\vec{v}_k, \vec{v}_k)$,} and $\langle \vec{\theta} \rangle$. (ii) \mbox{$\lim_{n \to \infty}\textrm{cov}\bigr(\vec{\theta}_{k}, \vec{\theta}_{k+n}\bigr) = 0$.} \mbox{(iii) $\lim_{n \to \infty}n\cdot\textrm{cov}\bigr(\vec{\theta}_{k}, \vec{\theta}_{k+n} - \vec{\theta}_{k+n+1}\bigr) = 0$.} For example, the latter two assumptions hold true if the weight correlation function decays as $\textrm{cov}\bigr(\vec{\theta}_{k}, \vec{\theta}_{k+n}\bigr) \propto n^{-2}$ or  faster. In the setup described in \cref{sec:var_main}, the weight correlations will even decay exponentially fast (see \cref{sec:appendix_var_tau_assumptions}).

We assume that that $\langle \Vec{\theta}\rangle$, $\vec{\Sigma_\theta}$ and $\vec{\Sigma_v}$ exist and are finite. Without loss of generality, let  $\langle \Vec{\theta}\rangle = 0$. We consider only the one-dimensional case. For the multidimensional case, simply apply the proof in the direction of any basis vector individually. Note, that the relation still holds if $[\vec{\Sigma}, \vec{\Sigma_v}] \neq 0$. In this case, $\sigma_{\theta, i}^2$ and $\sigma_{v, i}^2$ would just be the variances of the weight and the velocity in the given direction but no longer necessarily eigenvalues of $\vec{\Sigma}$ and $\vec{\Sigma_v}$.

The remaining two assumptions (ii) and (iii) of \cref{thm:correlation_time} can now be written as
\begin{align}
    \lim\limits_{m \rightarrow \infty}{\langle\theta_k \theta_{k+m} \rangle} &= 0 \label{eq:corr_time_proof_asumption_1}\\
    \lim\limits_{m \rightarrow \infty}{m\bigl(\langle\theta_k \theta_{k+m} \rangle - \langle\theta_k \theta_{k+m+1} \rangle\bigr)} &= 0 \label{eq:corr_time_proof_asumption_2} \ .
\end{align}

We begin the proof with the following chain of equations
\begin{align}
    \sigma_\theta^2 &= \left\langle \theta_{k}^2 \right \rangle \nonumber\\
    &= \left\langle \left( \theta_k - \theta_{k+J} + \theta_{k+J} \right)^2 \right \rangle \nonumber\\
    &= \left\langle \left( \theta_k - \theta_{k+J}\right)^2 \right \rangle -2\left\langle \theta_{k+J}^2 \right \rangle +2\left\langle \theta_k \theta_{k+J} \right \rangle + \left\langle \theta_{k+J}^2 \right \rangle \ , \label{eq:corr_time_proof_chain_1}
\end{align}
which holds for any integer $J$. We have $\left\langle \theta_{k+J}^2 \right \rangle = \left\langle \theta_{k}^2 \right \rangle$ since the expectation value cannot depend on $k$. Additionally, by  definition we have $v_k = \theta_k - \theta_{k-1}$ which yields \begin{align}
     \theta_k - \theta_{k+J} = \sum_{i=1}^J v_{k+i} \ .
 \end{align}
 Therefore, we can rewrite \cref{eq:corr_time_proof_chain_1} as follows
 \begin{align}
     2\sigma_\theta^2 &= 2\left\langle \theta_k \theta_{k+J} \right\rangle + \sum_{i,j = 1}^J \langle v_{k+i} v_{k+j} \rangle \nonumber\\
     &= 2\left\langle \theta_k \theta_{k+J} \right\rangle + \sum_{i,j = 1}^J \langle v_{k} v_{k+j-i} \rangle \nonumber\\
     &= 2\left\langle \theta_k \theta_{k+J} \right\rangle + \sum_{m = 0}^{J-1} \sum_{n = -m}^m \langle v_{k} v_{k+n} \rangle \ ,
 \end{align}
 where we first shifted the index within the expectation value and then restructured the sum by defining $m \coloneqq \max(i, j) -1$ and $n \coloneqq j-i$. We now take the limit of $J \rightarrow \infty$ and because of \cref{eq:corr_time_proof_asumption_1} and the assumption of a finite $\sigma_\theta^2$ we have
 \begin{align}
     \sum_{m = 0}^{\infty} \sum_{n = -m}^m \langle v_{k} v_{k+n} \rangle &< \infty\\
     \Rightarrow \sum_{n = -\infty}^\infty \langle v_{k} v_{k+n} \rangle &= 0  \ . \label{eq:corr_time_proof_limit_1}
 \end{align}
 
 We note that $\langle v_{k} v_{k+n} \rangle = \langle v_{k} v_{k-n} \rangle$ because we can shift the index,  and the two factors commute. Substituting this relation into \cref{eq:corr_time_proof_limit_1} yields
 \begin{align}
    \sum_{n = 1}^\infty \langle v_{k} v_{k+n} \rangle &=  -\frac{1}{2} \langle v_{k} v_{k} \rangle \nonumber\\
    &= -\frac{1}{2} \sigma_v^2  \ . \label{eq:corr_time_proof_main_1}
 \end{align}

 For the second part of the proof we will start again with $v_k = \theta_k - \theta_{k-1}$ and the following sum
 \begin{align}
     \sum_{n = 1}^m n\langle v_{k} v_{k+n} \rangle &= \sum_{n = 1}^m n \bigl(2\langle \theta_{k} \theta_{k+n} \rangle - \langle \theta_{k-1} \theta_{k+n} \rangle - \langle \theta_{k} \theta_{k+n-1} \rangle \bigr) \nonumber\\
     &= -\langle \theta_{k} \theta_{k} \rangle + \langle \theta_{k} \theta_{k+m} \rangle + m\bigl(\langle\theta_k \theta_{k+m} \rangle - \langle\theta_k \theta_{k+m+1} \rangle\bigr) \ ,
 \end{align}
 where nearly all terms  cancel  each other again due to the fact that we can shift the index within the expectation value. By taking the limit $m \rightarrow \infty$ and using the assumptions (ii) and (iii) (\cref{eq:corr_time_proof_asumption_1,eq:corr_time_proof_asumption_2}) we have
 \begin{align}
     \sum_{n = 1}^\infty n\langle v_{k} v_{k+n} \rangle = -\langle \theta_{k} \theta_{k} \rangle \ . \label{eq:corr_time_proof_main_2}
 \end{align}

 Finally, by dividing \cref{eq:corr_time_proof_main_2} by \cref{eq:corr_time_proof_main_1} we arrive at the final expression
 \begin{align}
     \frac{2 \sigma_{\theta, i}^2}{\sigma_{v, i}^2} = \frac{\sum_{n=1}^\infty n \, \langle v_{k} v_{k+n} \rangle}{\sum_{n=1}^\infty \langle v_{k} v_{k+n} \rangle} \ .
 \end{align}

%%%%%%%%%%%%%%%%%%%%%%%%%%%%%%%%%%%%%%%%%%%%%%%%%%
%%%%%%%%%%%%%%%%%%%%%%%%%%%%%%%%%%%%%%%%%%%%%%%%%%
%%%%%%%%%%%%%%%%%%%%%%%%%%%%%%%%%%%%%%%%%%%%%%%%%%
\FloatBarrier
\section{Calculation of the Noise Autocorrelation}
\label{sec:appendix_corr_calc}

We want to calculate the autocorrelation function of epoch-based SGD for a fixed weight vector $\vec{\theta}$ and under the assumption that the total number of examples is an integer multiple of the number of examples per batch. For that we repeat the following definitions:
\begin{align}
    \vec{\delta g}_k(\vec{\theta}) &\coloneqq \frac{1}{S} \sum_{n \in \mathcal{B}_k} \vec{\nabla}\bigl(l(\vec{\theta}, x_n) - L(\vec{\theta})\bigr)\\
    \mathcal{B}_k = \{n_1, ..., n_S\} &\dots \text{ batch of step $k$, sampling without replacement within epoch}\\
    n_j \in \{1, \dots, N\}\\
    N &\dots \text{ total number of examples}\\
    S &\dots \text{ number of examples per batch}
\end{align}

We can rewrite the noise terms as follows:
\begin{align}
    \vec{\delta g}_k(\vec{\theta}) &= \frac{1}{S} \sum_{n \in \mathcal{B}_k} \vec{\delta g_e}(n, \vec{\theta})\nonumber \\
    &= \frac{1}{S} \sum_{n = 1}^N \vec{\delta g_e}(n, \vec{\theta})s_k^n\\
    s_k^n &\coloneqq \mathbf{1}_{\mathcal{B}_k}(n)\nonumber\\ 
    &=\begin{cases} 1\ \text{ if }~ n \in \mathcal{B}_k \\ 0 \ \text{ if }~ n \notin \mathcal{B}_k \end{cases}\\
    \vec{\delta g_e}(n, \vec{\theta}) &\coloneqq \vec{\nabla}\bigl(l(\vec{\theta}, x_n) - L(\vec{\theta})\bigr) \ .
\end{align}

Let $h \geq 0$ be fixed. The correlation matrix can be expressed as
\begin{align}
    \textrm{cov}\bigl(\Vec{\delta g}_k(\Vec{\theta}), \Vec{\delta g}_{k+h}(\Vec{\theta})\bigr) &= \mathbb{E}\left[ \vec{\delta g}_k(\vec{\theta})\vec{\delta g}_{k+h}(\vec{\theta})^\top\right]\nonumber\\
    &= \frac{1}{S^2} \sum_{n, \Tilde{n} = 1}^N \vec{\delta g_e}(n, \vec{\theta})\vec{\delta g_e}(\Tilde{n}, \vec{\theta})^\top \mathbb{E}\left[s_k^ns_{k+h}^{\Tilde{n}}\right] \ .\label{eq:prob_cov_main}
\end{align}

The expectation value of $s_k^n = \mathbf{1}_{\mathcal{B}_k}(n)$ is the probability that example $n$ is part of batch $k$. Because every example is equally likely to appear in a given batch, this probability is equal to $S/N$.
\begin{align}
    \mathbb{E}\left[s_k^n\right] &= \textrm{P}\!\left(s_k^n = 1\right)\nonumber\\
    & = \frac{S}{N} \ .
\end{align}

Similarly we can calculate the desired correlation:
\begin{align}
    \mathbb{E}\left[s_k^ns_{k+h}^{\Tilde{n}}\right] &= \textrm{P}\!\left(s_k^n = 1, s_{k+h}^{\Tilde{n}} = 1 \right)\nonumber\\
    &= \textrm{P}\!\left(s_k^n = 1\right)\textrm{P}\!\left(s_{k+h}^{\Tilde{n}} = 1 \mid s_k^n = 1\right)\nonumber\\
    &= \frac{S}{N} \, \textrm{P}\!\left(s_{k+h}^{\Tilde{n}} = 1 \mid s_k^n = 1\right) \ .
\end{align}

The last term can be split up into different probabilities for different values of h. We can also distinguish the case where the two steps $k$ and $k+h$ are within the same epoch ($\textrm{ep}(k) = \textrm{ep}(k+h)$) or in different epochs ($\textrm{ep}(k) \neq \textrm{ep}(k+h)$).
\begin{align}
    \textrm{P}\!\left(s_{k+h}^{\Tilde{n}} = 1 \mid s_k^n = 1\right) &= \delta_{h,0} \cdot \textrm{P}\!\left(s_{k}^{\Tilde{n}} = 1 \mid s_k^n = 1\right) + (1 - \delta_{h,0}) \, \cdot \nonumber\\
    & \quad \Bigl[ \textrm{P}\bigl(\textrm{ep}(k) = \textrm{ep}(k+h) \bigr) \textrm{P}\!\left(s_{k+h}^{\Tilde{n}} = 1 \mid s_k^n = 1, \textrm{ep}(k) = \textrm{ep}(k+h) \right) + \nonumber \\
    & \quad \ \textrm{P}\bigl(\textrm{ep}(k) \neq \textrm{ep}(k+h) \bigr) \textrm{P}\!\left(s_{k+h}^{\Tilde{n}} = 1 \mid s_k^n = 1, \textrm{ep}(k) \neq \textrm{ep}(k+h) \right)  \Bigr] \ . \label{eq:prob_main}
\end{align}

The first term of the right hand side of \cref{eq:prob_main} is the probability that a given example occurs in a batch, assuming that we already know one of the examples of that batch.
\begin{align}
    \textrm{P}\!\left(s_{k}^{\Tilde{n}} = 1 \mid s_k^n = 1\right) &=  \textrm{P}\!\left(s_{k}^{n} = 1 \mid s_k^n = 1\right) \cdot \, \delta_{n,\Tilde{n}} + \textrm{P}\!\left(s_{k}^{\Tilde{n}} = 1 \mid s_k^n = 1, n \neq \Tilde{n} \right) \, (1 - \delta_{n,\Tilde{n}})\nonumber\\
    &= 1 \cdot \, \delta_{n,\Tilde{n}} + \frac{S-1}{N-1} \, (1 - \delta_{n,\Tilde{n}})\nonumber\\
    &= \frac{N-S}{N-1}\, \delta_{n,\Tilde{n}} + \textrm{const.}
\end{align}

The second term of \cref{eq:prob_main} is multiplied by $(1 - \delta_{h,0})$. Therefore, we assume $h\geq 1$ for the following argument. That is, we want to know the probabilities under the assumption that we are comparing examples from different batches. If the two batches are still from the same epoch, examples cannot repeat as the total number of examples is an integer multiple of the number of examples per batch and because of that every example is shown only once per epoch. Therefore, for $h\geq 1$ holds:
\begin{align}
    \textrm{P}\!\left(s_{k+h}^{\Tilde{n}} = 1 \mid s_k^n = 1, \textrm{ep}(k) = \textrm{ep}(k+h) \right) &= 0 \cdot \delta_{n,\Tilde{n}} + \frac{S}{N-1} \, (1 - \delta_{n,\Tilde{n}}) \nonumber\\
    &= -\frac{S}{N-1}\delta_{n,\Tilde{n}} + \textrm{const.}
\end{align}

If we consider batches from different epochs, the probability becomes independent of the given examples:
\begin{align}
    \textrm{P}\!\left(s_{k+h}^{\Tilde{n}} = 1 \mid s_k^n = 1, \textrm{ep}(k) \neq \textrm{ep}(k+h) \right) &= \frac{S}{N}\nonumber\\
    &= \textrm{const.}
\end{align}

Lastly, we need to know the probability that two given batches $k$ and $k+h$ are from the same epoch:
\begin{align}
    \textrm{P}\bigl(\textrm{ep}(k) = \textrm{ep}(k+h) \bigr) = \mathbf{1}_{\{1, \dots, M\}}(h)\,\frac{M - h}{M} \ ,
\end{align}
where $M=N/S$ is again the number of batches per epoch.

We can now combine all derived probabilities and arrive at the following relation:
\begin{align}
    \mathbb{E}\left[s_k^ns_{k+h}^{\Tilde{n}}\right] &= \delta_{n,\Tilde{n}}\,\frac{S}{N}\frac{N-S}{N-1}\left( \delta_{h,0} - \mathbf{1}_{\{1, \dots, M\}}(h)\frac{S}{N-S}\frac{M-h}{M}\right) + \textrm{const.}\nonumber\\
    &= \delta_{n,\Tilde{n}}\,S^2\left(\frac{1}{S} - \frac{1}{N}\right)\frac{}{}\frac{1}{N-1}\left( \delta_{h,0} - \mathbf{1}_{\{1, \dots, M\}}(h)\frac{M-h}{M\,(M-1)}\right) + \textrm{const.}
\end{align}

If we now also consider negative values for h, the expression depends only on the absolute value of h due to symmetry.

By using the following two helpful relations:
\begin{align}
    \sum_{n, \Tilde{n} = 1}^N \vec{\delta g_e}(n, \vec{\theta})\vec{\delta g_e}(\Tilde{n}, \vec{\theta})^\top \,\delta_{n,\Tilde{n}} &= \sum_{n = 1}^N \vec{\delta g_e}(n, \vec{\theta})\vec{\delta g_e}(n, \vec{\theta})^\top\nonumber\\
    &\eqqcolon (N-1)\,\vec{C}_0(\vec{\theta}) \ ,\\
    \sum_{n, \Tilde{n} = 1}^N \vec{\delta g_e}(n, \vec{\theta})\vec{\delta g_e}(\Tilde{n}, \vec{\theta})^\top \cdot 1 &= \left(\sum_{n = 1}^N \vec{\delta g_e}(n, \vec{\theta})\right) \left(\sum_{n = 1}^N \vec{\delta g_e}(n, \vec{\theta})^\top \right)\nonumber\\
    &=\vec{\nabla}\bigl(L(\vec{\theta}) - L(\vec{\theta})\bigr)\vec{\nabla}^\top\bigl(L(\vec{\theta}) - L(\vec{\theta})\bigr)\nonumber\\
    & = 0 \ ,
\end{align}
we can insert the expectation value $\mathbb{E}\left[s_k^ns_{k+h}^{\Tilde{n}}\right]$ into \cref{eq:prob_cov_main} and arrive at the final expression:
\begin{align}
    \textrm{cov}\bigl[\Vec{\delta g}_k(\Vec{\theta}), \Vec{\delta g}_{k+h}(\Vec{\theta})\bigr] &= \textrm{cov}\bigl[\Vec{\delta g}_k(\Vec{\theta}), \Vec{\delta g}_k (\Vec{\theta})\bigr]\cdot \left( \delta_{h,0} - \mathbf{1}_{\{1, ..., M\}}(|h|)\frac{M-|h|}{M(M-1)} \right),\\
    \textrm{cov}\bigl[\Vec{\delta g}_k(\Vec{\theta}), \Vec{\delta g}_k (\Vec{\theta})\bigr] &= \left(\frac{1}{S} - \frac{1}{N}\right) \vec{C}_0(\vec{\theta}) \ .
\end{align}

\FloatBarrier
\section{Anti-Correlations for fluctuating Weights}
\label{sec:appendix_non_static_weights}

In section \cref{sec:correlation_main}, we described a straightforward theoretical argument explaining the anti-correlations observed in the gradient noise of SGD. However, one simplifying assumption in that section was that of a static weight vector. In practice, the weights evolve during training. In this section, we derive predictions for the anti-correlations in the gradient noise and for the weight variances near a minimum of the loss, taking into account the fact that weights are fluctuating during training.

The main tool to derive predictions for this setting is to involve not only a second-order Taylor approximation of the full loss function $L(\vec{\theta})$, but also of all the individual minibatch loss functions 
\begin{align}
    l_{\mathcal{B}_k}(\vec{\theta}) \coloneqq \frac{1}{S} \sum_{n\in\mathcal{B}_k} l(\vec{\theta}, x_n)
\end{align}
with the minibatch size $S$. This will allow us to predict anti-correlations in the noise and lower-than-expected weight variances in relatively flat directions, without the assumption of static weights.

First, we present a definition of the extended model and a derivation of how the anti-correlations in the noise can still emerge. In the subsection after that, we derive predictions for the weight variances in this extended model.

\subsection{Extended Model and Anti-Correlated Noise}

Assume that the fluctuating weights $\vec{\theta}_k$ are close to a minimum $\vec{\theta}^*$ and that they stay close enough to that minimum that we can approximate the full loss and the minibatch loss functions as quadratic functions. Without loss of generality, we assume $\vec{\theta}^* = 0$, which gives us
\begin{align}
    L(\vec{\theta}) \approx \vec{\theta}^\top \vec{H} \vec{\theta} + L(\vec{\theta}^*)
\end{align}
and
\begin{align}
    l_{\mathcal{B}_k}(\vec{\theta}) \approx \vec{\theta}^\top\vec{g}_k^* + \vec{\theta}^\top (\vec{H} + \vec{\delta H}_k) \vec{\theta} + L(\vec{\theta}^*) \ .
\end{align}
The vector $\vec{g}_k^*$ and the matrix $\vec{\delta H}_k$ are defined as the differences between the respective full loss function and the minibatch loss function quantities, with $\vec{g}_k^* \coloneqq \vec{\nabla}l_{\mathcal{B}_k}(\vec{\theta}^*)$ and $\vec{\delta H}_k \coloneqq \vec{\nabla}\vec{\nabla}^\top l_{\mathcal{B}_k}(\vec{\theta}^*) - \vec{H}$.

Since the average of the minibatch gradients at the minimum over one epoch of example selection without replacement must be exactly zero, by the same argument as in \cref{sec:correlation_main}, we can assume that the $\vec{g}_k^*$ are anti-correlated,
\begin{align}
    {\rm cov}(\vec{g}_k^*, \vec{g}_{k+h}^*) = \begin{cases}
        \vec{C}^* \, , & h=0\\
        -\frac{M - |h|}{M(M-1)}\vec{C}^* \, ,&1\leq|h|\leq M \\
        0 \, , &|h|>M
    \end{cases} \ .
\end{align}
Here, we define $\vec{C}^* \coloneqq {\rm cov}(\vec{g}_k^*, \vec{g}_{k}^*)$. Notice that we can make this assumption without any restriction on the weights being static, since these gradients are independent of $\vec{\theta}_k$ and depend only on the minimum $\vec{\theta}^*$. Following this argument, the $\vec{\delta H}_k$ would be anti-correlated as well, but we neglect these correlations and assume that they are not correlated over time. Numerical investigations of this model show the same behavior as derived later in this section, independent of whether the Hessian noise terms $\vec{\delta H}_k$ are anti-correlated or not. We further assume that the elements of $\vec{\delta H}_k$ are not correlated with the elements of $\vec{g}_k^*$.

With these definitions, we can now write down explicitly the gradient noise term -- the difference between the full loss and the minibatch loss gradient -- and its weight dependence,
\begin{align}
    \vec{\delta g}_k = \vec{g}_k^* + \vec{\delta H}_k \vec{\theta}_k \ .
\end{align}
We can now consider the correlations of this gradient noise. With the assumption of uncorrelated Hessian noise terms, it is straightforward to derive the following relation,
\begin{align}
    {\rm cov}(\vec{\delta g}_k , \vec{\delta g}_{k+h}) = \begin{cases}
        \vec{C} \, , & h=0\\
        -\frac{M - |h|}{M(M-1)}\vec{C}^* \, ,&1\leq|h|\leq M \\
        0 \, , &|h|>M
    \end{cases} 
\end{align}
with the covariance of the noise in this model being
\begin{align}
    \vec{C} \coloneqq\vec{C}^*  + \langle\vec{\delta H}_k \vec{\theta}_k\vec{\theta}_k^\top\vec{\delta H}_k \rangle \ .
    \label{eq:appendix_noise_covariance_extended}
\end{align}
If the second term on the right-hand side of the above equation is negligible compared to the first term, we would have recovered the anti-correlations described in section \cref{sec:correlation_main}. In the next subsection, we will see that when we assume the variances of the elements of $\vec{\delta H}$ follow predictions for an actual neural network optimization, we can indeed recover the anti-correlations approximately.

\subsection{Weight Variances in the extended model}

To solve for the weight variances in this extended model, we first need to consider what to assume for the variances of the entries of $\vec{g}_{k}^*$ and $\vec{\delta H}_k$. For that, we will essentially consider a network with a one-dimensional output employing the mean squared error. However, we expect similar results for the case of a multidimensional output employing the softmax function with cross-entropy used as the loss.

Let $y_n$ be the desired output given by the dataset for an example $x_n$, and let $f_n(\vec{\theta})$ be the network output. The full loss is then given by
\begin{align}
    L(\vec{\theta}) = \frac{1}{N} \sum_{n=1}^N \frac{1}{2}(f_n(\vec{\theta}) - y_n)^2 \ .
\end{align}
Assume the weights are close to a minimum $\vec{\theta}^*$ of that loss and that we can linearize the network $f_n(\vec{\theta})$ near that minimum. Then we have
\begin{align}
    f_n(\vec{\theta}) \approx f_n(\vec{\theta}^*) + (\vec{\theta} - \vec{\theta}^*)^\top \vec{\phi}_n
\end{align}
with $\vec{\phi}_n = \vec{\nabla}f_n(\vec{\theta}^*)$. This linearization of the network output is closely related to the neural tangent kernel concept \cite{Jacot.2018}. Without loss of generality, we again set $\vec{\theta}^* = 0$, which allows us to write the gradient of the minibatch loss functions at the minimum as
\begin{align}
    \vec{g}_k^* = \frac{1}{S} \sum_{n \in \mathcal{B}_k} \vec{\phi}_n \delta y_n
\end{align}
with $\delta y_n \coloneqq f_n(\vec{\theta}^*) - y_n$. The Hessian in this framework is given by $\vec{H} = \frac{1}{N}\sum_{n=1}^N \vec{\phi}_n\vec{\phi}_n^\top$. Therefore, under the assumption that the $\delta y_n$ are not correlated with the network gradients $\vec{\phi}_n$, we get proportionality between the Hessian and the covariance of the minibatch gradients at the minimum in the model. It is straightforward to show that
\begin{align}
    \vec{C}^* \equiv \vec{H}\left(\frac{1}{S} - \frac{1}{N}\right)\frac{1}{N}\sum_{n=1}^N \delta y_n^2
\end{align}
giving us that this noise not only commutes with the Hessian in this model, but is even proportional to it.

\begin{figure}[ht]
    \centering
    \includegraphics[]{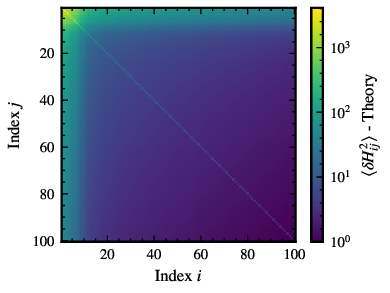}
    \includegraphics[]{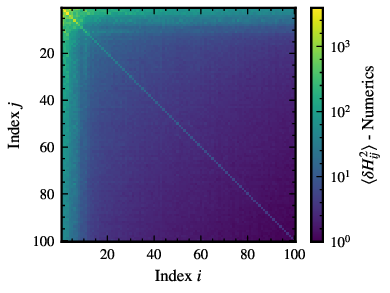}
    \caption{Theoretical prediction (left) and numerical estimates (right) of the variance of Hessian noise terms, $\langle \delta H_{ij}^2 \rangle$. Noise variances were computed at the weight vector obtained after 300 epochs of initial training. The dataset was split into $M=1000$ batches of $S=50$ examples each, and for each batch, the deviation of Hessian elements from the full-batch Hessian was evaluated in the subspace spanned by the top 100 Hessian eigenvectors, i.e., $\delta H_{ij}^{(k)}(\vec{\theta}_K)$ for $k = K, \dots, K+M$.}
    \label{fig:appendix_hessian_noise_terms}
\end{figure}

The Hessian noise terms are given by 
\begin{align}
    \vec{\delta H}_k = \frac{1}{S}\sum_{n \in \mathcal{B}_k} \vec{\phi}_n\vec{\phi}_n^\top - \vec{H} \ .
\end{align}
Assume, without loss of generality, that the Hessian is diagonal in the basis we are considering, giving us $H_{ij} = \lambda_i\delta_{ij}$ with the Kronecker delta $\delta_{ij}$. For a high number of examples $N$, assuming that the network gradients $\vec{\phi}_n$ are approximately Gaussian distributed, and by employing Wick's theorem, it is straightforward to derive the following relation for the covariance of the elements of the Hessian noise,
\begin{align}
    {\rm cov}(\delta H_{ij}, \delta H_{kl}) \approx \frac{1}{S}\lambda_i\lambda_j\left( \delta_{ik}\delta_{jl} + \delta_{il}\delta_{jk}\right) \ .
\end{align}
Numerically evaluated Hessian noise terms of the LeNet model at the end of training indeed follow this prediction, as shown in \cref{fig:appendix_hessian_noise_terms}.

With the variances of the noise terms now determined, we will derive the weight variance predictions for our extended setup incorporating the more general anti-correlated noise, taking into account the fluctuating weight vector. We will derive the relations without momentum -- $\beta = 0$ -- but later give numerically validated relations with momentum. The derivation is very similar to what we described before in \cref{sec:var_main}.

We begin with the update equation for SGD in our extended model,
\begin{align}
    \vec{\theta}_{k+1} = \left(\vec{1} - \eta \vec{H} - \eta \vec{\delta H}_k\right) \vec{\theta}_k - \eta \vec{g}_k^* \ .
    \label{eq:appendix_extended_master}
\end{align}
By squaring both sides of the above equation, averaging over an infinite run of SGD optimization, and noticing that all relevant matrices commute under our assumptions of this subsection, we get
\begin{align}
    \vec{\Sigma} =& \vec{\Sigma} -2\eta\vec{H}\vec{\Sigma} + \eta^2\vec{H}^2\vec{\Sigma} + \eta^2\left\langle \vec{\delta H}_k \vec{\theta}_k\vec{\theta}_k^\top \vec{\delta H}_k \right\rangle \nonumber \\ 
    &+ \eta^2 \vec{C}^* + \eta \left[(1 - \eta \vec{H})\langle\vec{\theta}_k\vec{g}_k^{*\top}\rangle + \langle\vec{g}_k^*\vec{\theta}_k^\top\rangle(1 - \eta \vec{H})\right] \ .
    \label{eq:appendix_extended_var01}
\end{align}
We first analyze the second line of the above equation, which holds the term $\langle\vec{g}_k^*\vec{\theta}_k^\top\rangle$ and its transpose -- exactly the terms which incorporate the anti-correlations of the noise and which would be zero if there were no anti-correlations, just as in the previous derivation described in \cref{sec:var_main}.

An iteration of the update equation \cref{eq:appendix_extended_master} gives us
\begin{align}
    \vec{\theta}_k =& \sum_{m=1}^M \prod_{h=1}^{m-1}\left(\vec{1} - \eta\vec{H} - \eta \vec{\delta H}_{k-h} \right) \vec{g}_{k-m}^* \nonumber \\
    &+ \prod_{h=1}^M\left(\vec{1} - \eta\vec{H} - \eta \vec{\delta H}_{k-h} \right) \vec{\theta}_{k-M} \ .
\end{align}
Since we assume no correlations between the Hessian noise terms $\vec{\delta H}_k$ of different update steps and no correlation with the minimum gradient noise terms $\vec{g}_{k}^*$, the calculation of the correlation term $\langle\vec{g}_k^*\vec{\theta}_k^\top\rangle$ is very similar to the calculation of the corresponding term in \cref{sec:appendix_var_exact}. It is straightforward but lengthy to show that, in a basis where the Hessian is diagonal, the diagonal of the second line of \cref{eq:appendix_extended_var01} is given by
\begin{align}
    {\rm diag}\left(\eta^2 \vec{C}^* + \eta \left[(1 - \eta \vec{H})\langle\vec{\theta}_k\vec{g}_k^{*\top}\rangle + \langle\vec{g}_k^*\vec{\theta}_k^\top\rangle(1 - \eta \vec{H})\right]\right) = \eta^2\left(\eta\vec{H}\vec{\tau}\right)\vec{\sigma}_{g^*}^2
\end{align}
with $\vec{\sigma}_{g^*}^2 \coloneqq {\rm diag}(\vec{C}^*)$, and where the matrix $\vec{\tau}$ is diagonal as well, with the same correlation times $\tau_i$ on its diagonal we already described in \cref{sec:var_main}. The correlation times are approximately given by $\tau_i \approx \frac{1 + \beta}{\eta \lambda_i}$ for $\lambda_i > \lambda_{\rm cross} = \frac{3(1 - \beta)}{\eta M}$, and $\tau_i \approx \frac{M}{3}\frac{1 + \beta}{1 - \beta}$ for $\lambda_i < \lambda_{\rm cross}$, under the assumption $M(1 - \beta) \gg 1$ for the case with momentum.

Considering the first line on the right-hand side of \cref{eq:appendix_extended_var01}, we need to evaluate the term $\left\langle \vec{\delta H}_k \vec{\theta}_k\vec{\theta}_k^\top \vec{\delta H}_k \right\rangle$. With our assumption for the covariances of the Hessian noise terms, it is straightforward to derive that
\begin{align}
    \left\langle \vec{\delta H}_k \vec{\theta}_k\vec{\theta}_k^\top \vec{\delta H}_k \right\rangle = \frac{1}{S}\vec{H}\vec{\Sigma}\vec{H} + \frac{1}{S}\vec{H}{\rm Tr}(\vec{H}\vec{\Sigma}) \ . 
    \label{eq:appendix_hessian_noise_terms_and_weights}
\end{align}
The above evaluation is again a diagonal matrix with its diagonal given by
\begin{align}
    {\rm diag}\left(\left\langle \vec{\delta H}_k \vec{\theta}_k\vec{\theta}_k^\top \vec{\delta H}_k \right\rangle\right) = \frac{1}{S} \vec{H}^2 \vec{\sigma}_\theta^2 + \frac{1}{S}\vec{\lambda}\vec{\lambda}^\top \vec{\sigma}_\theta^2
\end{align}
with $\vec{\sigma}_\theta^2 = {\rm diag}(\vec{\Sigma})$ in the basis where the Hessian is diagonal, and $\vec{\lambda} = {\rm diag}(\vec{H})$.

At this point, we can also again consider the noise covariance discussed in the previous section in \cref{eq:appendix_noise_covariance_extended}. There, we had that if $\vec{C}^* + \langle \vec{\delta H}_k \vec{\theta}_k\vec{\theta}_k^\top \vec{\delta H}_k \rangle \approx \vec{C}^*$, the anti-correlations previously predicted by the static weight vector argument would still hold approximately. We assume that $\vec{C}^*$ and $\vec{H}$ commute and that $\vec{\Sigma} \preceq \frac{\eta}{2}\vec{H}^{-1}\vec{C}^*$, which essentially means that we assume the weight variance to be smaller than or equal to the prediction without any anti-correlations in the noise. This is true for the model discussed in the main text and, as we will see later in this subsection, this is also true for this extended model. Inserting this assumption into \cref{eq:appendix_hessian_noise_terms_and_weights} we get
\begin{align}
    \langle \vec{\delta H}_k \vec{\theta}_k\vec{\theta}_k^\top \vec{\delta H}_k \rangle \preceq \frac{1}{2S}\eta\vec{H}\left[\vec{C}^* + \vec{1}\,{\rm Tr}(\vec{C}^*)\right] \ .
\end{align}
With this approximation, we can see that this term is indeed negligible compared to $\vec{C}^*$. We assume that $\vec{C}^* \propto \vec{H}$ and that the Hessian has mostly small eigenvalues, therefore also meaning that ${\rm Tr}(\vec{C}^*)$ is at most an order of ten to a hundred times larger than its largest eigenvalue. Since we also assume that $\eta\lambda_i \ll 1$ for all eigenvalues, it is straightforward to see that any eigenvalue of $\frac{1}{2S}\eta\vec{H}\left[\vec{C}^* + \vec{1}\,{\rm Tr}(\vec{C}^*)\right]$ is small compared to the corresponding eigenvalue of $\vec{C}^*$ for the same eigenvector. Together, the anti-correlations of the noise in this extended model are approximately the same as discussed in the main text. However, even small deviations might lead to different predictions for the weight variances, especially in directions with small Hessian eigenvalues. Therefore, we need to further examine them in this extended model.

Having evaluated all the terms on the right-hand side of \cref{eq:appendix_extended_var01}, we can now rearrange the matrix equation and rewrite it as an equation for the diagonal of the respective matrices, giving
\begin{align}
    \left(2 \eta \vec{H} - \frac{1}{S}\eta^2\vec{\lambda}\vec{\lambda}^\top\right) \vec{\sigma}_\theta^2 \approx \eta^2(\eta\vec{H}\vec{\tau})\vec{\sigma}_{g^*}^2 \ .
\end{align}
Here, we neglected terms of order $\eta^2\lambda^2$, since we assume $\eta \lambda \ll 1$. By inverting the matrix prefactor on the left-hand side via the Sherman–Morrison formula for computing the inverse of a rank-1 update to a matrix, we can derive the weight variances in the respective eigendirections of the Hessian matrix. Incorporating momentum, the numerically validated prediction for the weight variance in this extended model is given by
\begin{align}
    \sigma_{\theta, i}^2 \approx \frac{1}{2}\tau_i\frac{\eta^2\sigma_{g^*, i}^2}{(1+\beta)(1-\beta)} + \frac{1}{4S(1+\beta)(1-\beta)^2}\sum_{j=1}^d \eta\lambda_j\tau_j\eta^2\sigma_{g^*, j}^2 \ . 
    \label{eq:appendix_extended_var02}
\end{align}
The first term on the right-hand side is exactly the same as in the model without the Hessian noise terms and, therefore, it does not reflect any changes to the anti-correlated noise due to a fluctuating weight vector. This term still predicts lower-than-expected weight variances for directions with small Hessian eigenvalues, $\lambda_i < \lambda_{\rm cross}$. However, the second term is a fixed value that is the same for every direction in weight space. Due to an additional factor of $\eta\lambda_j$, this second term is relatively small.

With the assumption that $\sigma_{g^*, i}^2 \propto \lambda_i$ and that $\eta\lambda_i \ll 1$, one can derive that for Hessian eigenvalues smaller than a minimal Hessian eigenvalue $\lambda_{\rm min}$, the prediction of our extended model for the weight variances deviates from the predictions given in the main text. This deviation occurs approximately when both terms on the right-hand side of \cref{eq:appendix_extended_var02} are equal. The sum in the fixed term essentially only depends on the directions of the largest Hessian eigenvalues, which are larger than $\lambda_{\rm cross}$. Terms for smaller $\lambda_j$ are suppressed by the correlation time $\tau_j = \frac{M(1+\beta)}{3(1-\beta)}$ being significantly smaller than $\frac{1+\beta}{\eta \lambda_j}$. It is straightforward to show that with this approximation, both terms on the right-hand side of \cref{eq:appendix_extended_var02} are equal for
\begin{align}
    \lambda_{\rm min} \approx \frac{3}{2N}\sum_{\lambda_i > \lambda_{\rm cross}}\lambda_i \ .
\end{align}
Furthermore, the corresponding minimal weight variance is then given by
\begin{align}
    \sigma_{\theta, {\rm min}}^2 = \sigma_{\theta, {\rm max}}^2 \frac{1}{2S(1-\beta)} \sum_{i > i_{\rm cross}} \eta \lambda_i \ .
\end{align}

With these predictions, we can now assess whether the modifications to the anti-correlations of the noise due to a fluctuating weight vector change the weight variance. If we consider the LeNet network examined in the main text, the sum over the largest Hessian eigenvalues is on the order of about 10 to 20 times the largest Hessian eigenvalue, meaning that $\lambda_{\rm min} \approx 10^{-4}\lambda_{\rm max}$ for $N = 50000$, which is smaller than the smallest Hessian eigenvalue among the 5000 largest examined. With the hyperparameters employed, at the end of the LeNet network training schedule, the lower bound for the weight variance $\sigma_{\theta, {\rm min}}^2$ given by this extended model is about $10^{-2}\sigma_{\theta, {\rm max}}^2$. So even though this extended model does not predict the weight variances to become arbitrarily small for smaller and smaller Hessian eigenvalues, it still predicts a significantly lower weight variance than the constant weight variance prediction of $\sigma_{\theta, i}^2 = \sigma_{\theta, {\rm max}}^2$ for all directions. This constant variance prediction was the previous prediction in the literature without consideration of anti-correlations in the noise.

%%%%%%%%%%%%%%%%%%%%%%%%%%%%%%%%%%%%%%%%%%%%%%%%%%
%%%%%%%%%%%%%%%%%%%%%%%%%%%%%%%%%%%%%%%%%%%%%%%%%%
%%%%%%%%%%%%%%%%%%%%%%%%%%%%%%%%%%%%%%%%%%%%%%%%%%
\section{Effect of uneven Batch size on Anti-Correlated Noise}
\label{sec:appendix_non_integer_M}

As discussed in \cref{sec:correlation_main}, the derivation of the noise correlation formula \cref{eq:noise_corr} requires that the number of batches per epoch $M = N/S$ be an integer. This ensures that each example is seen exactly once per epoch when sampling without replacement, and results in exact cancellation of noise contributions over one epoch.

In practical training scenarios, however, $N$ might not be an integer multiple of $S$. In this section, we address the implications of this more general case and explain why the conclusions of our model remain valid, albeit approximately, even when $N/S \notin \mathbb{N}$. Furthermore, we quantify the residual noise that remains due to the incomplete final batch and compare this to the typical noise scale when sampling with replacement.

When the total number of examples $N$ is not divisible by the batch size $S$, a standard epoch is composed of $\lfloor N/S \rfloor$ full batches and a final partial batch containing the remaining $L \coloneqq N - S \lfloor N/S \rfloor < S$ examples. This results in a total of $M = \lceil N/S \rceil$ batches per epoch. The final batch introduces an asymmetry, as its noise contribution cannot fully cancel with the rest of the epoch, thereby breaking the exact noise cancellation assumed in \cref{eq:noise_corr}.

To illustrate the quantitative impact, we compare the variance of the mean noise over one epoch in three different scenarios:

\begin{enumerate}
    \item Sampling with replacement (no anti-correlations):
    \begin{align}
        \delta g_k^{(\text{w. r.})} &= \frac{1}{S} \sum_{n \in \mathcal{B}_k^{(\text{w. r.})}}\delta g_n^{(e)} \ , \\
        \left\langle \left( \frac{1}{M} \sum_{k=1}^M \delta g_k^{(\text{w. r.})} \right)^2 \right\rangle &= \frac{1}{N} \sigma^2_{\delta g, e} \ ,
    \end{align}
    where $\delta g_n^{(e)} = \nabla_{\theta}(l({\theta}, x_n) - L({\theta}))$ denotes the individual example noise. Here, $\langle\dots\rangle$ denotes an average over all possible batches of examples sampled with replacement.

    \item Sampling without replacement and $N/S \in \mathbb{N}$ (perfect anti-correlations):
    \begin{align}
        \left( \frac{1}{M} \sum_{k=1}^M \delta g_k^{(\text{w/o r.})} \right)^2  = 0 \ .
    \end{align}
    Notice that in this case, the noise term sum is exactly zero for any valid without replacement example batch composition, and not only when averaged.

    \item Sampling without replacement and $N/S \notin \mathbb{N}$ (imperfect anti-correlations):
    The final batch of size $L < S$ contributes unequally:
    \begin{align}
        \delta g_M^{(\text{w/o r.})} &= \frac{1}{L} \sum_{n \in \mathcal{B}_M^{(\text{w/o r.})}}\delta g_n^{(e)} \ ,
    \end{align}
    leading to a residual variance:
    \begin{align}
        \left\langle \left( \frac{1}{M} \sum_{k=1}^M \delta g_k^{(\text{w/o r.})} \right)^2 \right\rangle &= \left\langle \left( \frac{1}{M} \left( \frac{1}{L} - \frac{1}{S} \right) \sum_{n \in \mathcal{B}_M^{(\text{w/o r.})}} \delta g_n^{(e)} \right)^2 \right\rangle \\
        &\lesssim \frac{1}{M} \left( \frac{S}{L} - 1 \right) \frac{1}{N} \sigma^2_{\delta g, e} \ .
    \end{align}
    Here, $\langle\dots\rangle$ denotes an average over all possible batches of examples sampled without replacement.
\end{enumerate}

This implies that, while the anti-correlations no longer cancel the noise exactly in the non-integer case, the remaining variance is still significantly reduced -- by roughly a factor of $\frac{1}{M}(\frac{S}{L} - 1)$ compared to the variance in the sampling-with-replacement scenario. For typical values in our experiments, such as $S/L \approx 2$ and $M \approx 1000$, this corresponds to a noise variance reduction by approximately three orders of magnitude. This residual variance introduces a non-zero lower bound on the weight variances, even for directions associated with small Hessian eigenvalues. Nonetheless, this bound is orders of magnitude smaller than the maximum variance along large-eigenvalue directions.

Although the derivation of \cref{eq:noise_corr} strictly assumes $N/S \in \mathbb{N}$, the theoretical insights and predictions remain robust for non-integer $N/S$. The primary mathematical challenge in generalizing the derivation lies in accounting for the unequal batch size in the final update step of each epoch. This asymmetry prevents exact noise cancellation, but the residual error is small enough to leave the main conclusions of our variance analysis intact. Nevertheless, for theoretical clarity and to fully exploit the noise cancellation effect, choosing a batch size $S$ such that $N/S$ is an integer is advisable when feasible.

%%%%%%%%%%%%%%%%%%%%%%%%%%%%%%%%%%%%%%%%%%%%%%%%%%
%%%%%%%%%%%%%%%%%%%%%%%%%%%%%%%%%%%%%%%%%%%%%%%%%%
%%%%%%%%%%%%%%%%%%%%%%%%%%%%%%%%%%%%%%%%%%%%%%%%%%
\FloatBarrier
\section{Extended Numerics}

\subsection{Training Schedule of LeNet}
\label{sec:appendix_loss_evolution}

In order to corroborate our theoretical findings, we have conducted a small-scale experiment. We have trained a LeNet architecture, similar to the one described in \cite{Feng.2021}, using the CIFAR10 dataset \cite{Krizhevsky.2009}. LeNet is a compact convolutional network comprised of two convolutional layers followed by three dense layers. The network comprises approximately 137,000 parameters. As our loss function, we employed Cross Entropy, along with an L2 regularization with a prefactor of $5\cdot10^{-3}$. We used SGD to train the network for 300 epochs, employing an exponential learning rate schedule that reduces the learning rate by a factor of $0.98$ each epoch. The initial learning rate is set at $5\cdot10^{-3}$, which eventually reduces to approximately $1.1\cdot10^{-5}$ after 300 epochs. The momentum parameter and the minibatch size $S$ are set to $0.9$ and $50$, respectively, which results in a thousand minibatches per epoch,  $M = 1000$. This hyperparameter combination achieved one of the top test accuracies in a small grid search. The setup achieves 95\% training accuracy and about 70\% testing accuracy. The evolution of loss and accuracy during training can be seen in \cref{{fig:numerics_training_lenet}}.

\begin{figure}[ht]
    \centering
    \includegraphics[]{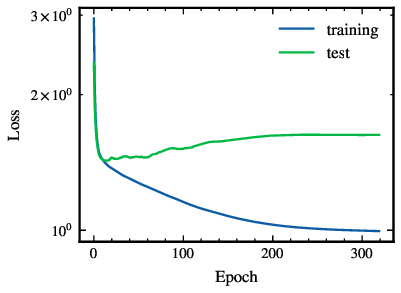}
    \includegraphics[]{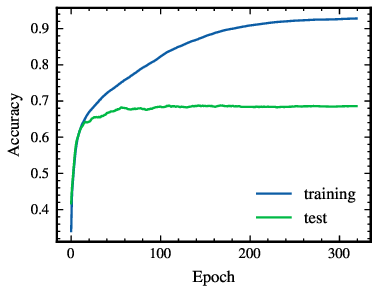}
    \caption{The evolution of the loss (left) and accuracy (right) during training of LeNet described in the main text. The statistics are shown for both training and test set. For the first 300 epochs, the exponential learning rate decay was used, and for the last 20 epochs, the learning rate was fixed at the final value of the exponential decay.}
    \label{fig:numerics_training_lenet}
\end{figure}

\begin{figure}[ht]
    \centering
    \includegraphics[]{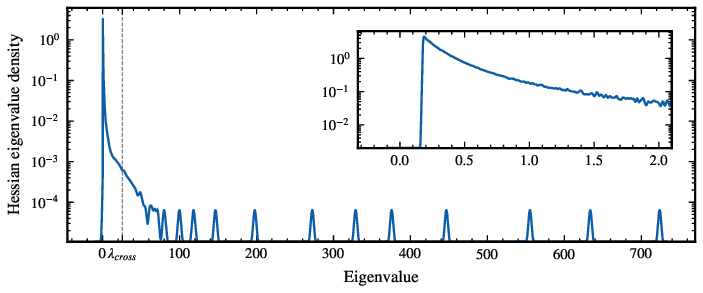}
    \caption{The distribution of the approximated 5,000 Hessian eigenvalues of the LeNet discussed in the main text. The inset shows that the smallest approximated eigenvalue has a magnitude of about 0.2.}
    \label{fig:hessian_density}
\end{figure}

Immediately after the initial schedule follows the analysis period of 20 additional epochs, corresponding to 20,000 update steps. We restrict our analysis to a specific subspace for which we approximate the 5,000 largest eigenvalues and their associated eigenvectors of the Hessian matrix at the beginning of the analysis period. The distribution of the approximated 5,000 eigenvalues is illustrated in \cref{fig:hessian_density}.

%%%%%%%%%%%%%%%%%%%%%%%%%%%%%%%%%%%%%%%%%%%%%%%%%%
%%%%%%%%%%%%%%%%%%%%%%%%%%%%%%%%%%%%%%%%%%%%%%%%%%
%%%%%%%%%%%%%%%%%%%%%%%%%%%%%%%%%%%%%%%%%%%%%%%%%%
\FloatBarrier
\subsection{Comparison with principal component analysis}
\label{sec:appendix_feng_tu}

Our approach to analysis sets itself apart from that of Feng \& Tu~\cite{Feng.2021} principally in the selection of the basis $\{\Vec{p_i}, \, i=1,\dots,d\}$ used for examining the weights. While they employ the principal components of the weight series - the eigenvectors of $\Vec{\Sigma}$ - we use the eigenvectors of the Hessian matrix $\Vec{H}(\Vec{\theta_K})$ computed at the beginning of the analysis period.

This choice enables us to directly create plots of variances and correlation time against the Hessian eigenvalue for each corresponding direction. Feng \& Tu devised a landscape-dependent flatness parameter $F_i$ for every direction $\Vec{p_i}$. However, with the assistance of the second derivative $F_i \approx \left( \partial^2 L(\Vec{\theta})/\partial \theta_i^2 \right)^{-\frac{1}{2}}$, where $\theta_i = \Vec{\theta}\cdot\Vec{p_i}$, this parameter can be approximated, provided this second derivative retains a sufficiently positive value. Hence, in the eigenbasis of the Hessian matrix, the flatness parameter can be approximated as $F_i \approx \lambda_i^{-\frac{1}{2}}$, facilitating comparability between our analysis and that of Feng \& Tu.

The principal component basis, as used by Feng \& Tu, holds a distinct advantage in their unregularized setup without weight decay. The advantage is connected to a near-linear drift of the weights they observed, even toward the end of training, in a direction roughly proportional to the overall weight vector. This behavior is a known instability in networks employing ReLU activations and softmax output layers: once 100\% training accuracy is achieved, scaling all parameters simply scales the logits, thereby increasing the confidence of correct predictions without affecting classification outcomes. Applying sufficient regularization through weight decay eliminates this drift.

To ensure comparability with Feng \& Tu, we omit weight decay in this appendix section. As a result, the weights exhibit a near-linear drift over time, similar to their observations. To isolate the stationary components of the dynamics, we subtract the mean velocity from both the weights and updates, redefining $\Vec{\theta}_k$ and $\Vec{v}_k$ as $\Vec{\theta}_k^{\textrm{(s)}} \coloneqq \Vec{\theta}_k - \Vec{\bar v}\cdot k$ and $\Vec{v}_k^{\textrm{(s)}} \coloneqq \Vec{v}_k - \Vec{\bar v}$, respectively. Variances are then computed for these adjusted quantities, $\Vec{\theta}_k^{\textrm{(s)}}$ and $\Vec{v}_k^{\textrm{(s)}}$, which show more stationary behavior. However, in Feng \& Tu's analysis, this linear movement is automatically subsumed in the first principal component due to its pronounced variance. Hence, there's no necessity for additional subtraction of this drift in the weight covariance eigenbasis.

Yet, the weight covariance eigenbasis has a significant shortcoming: it yields artifacts. This is because $\Vec{\Sigma}$ is calculated as an average over a finite data set, skewing its eigenvalues from the anticipated distribution. Consequently, the resultant eigenvectors may not align perfectly with the expected ones. This issue is further exacerbated due to the high dimensionality of the underlying space. 

\begin{figure}[ht]
    \centering
    \includegraphics{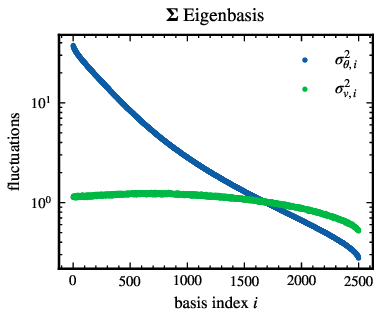}
    \includegraphics{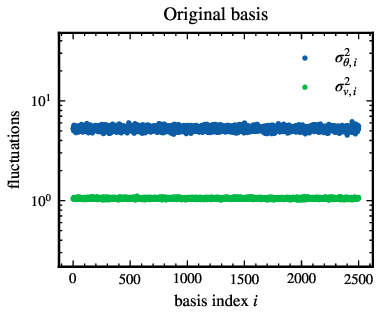}
    \\
    \includegraphics{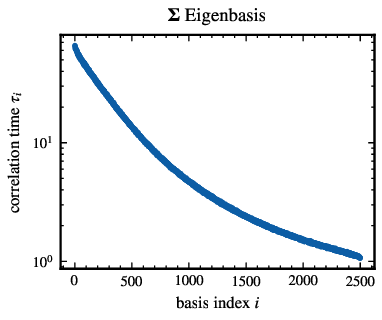}
    \includegraphics{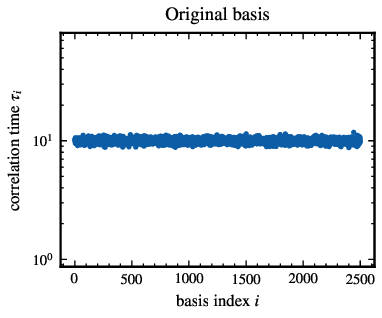}
    \caption{Comparison of weight and velocity fluctuations for synthetic data analyzed in two different bases. We define the variance in weight, $\sigma_{\theta, i}^2$, as $\Vec{p_i}^\top\Vec{\Sigma}\Vec{p_i}$, and the variance in velocity, $\sigma_{v, i}^2$, as $\Vec{p_i}^\top\Vec{\Sigma_v}\Vec{p_i}$. The correlation time, $\tau_i$, is given by $2 \sigma_{\theta, i}^2 / \sigma_{v, i}^2$. The synthetic data was generated by simulating SGD for 12,000 steps within a 2,500-dimensional space featuring an isotropic quadratic potential and isotropic noise. In the original basis analysis, both the variance and correlation time, as expected, retain isotropy. However, when the analysis is conducted in the eigenbasis of the weight covariance matrix, a pronounced anisotropy emerges.
    }
    \label{fig:feng_basis_comp_mock}
\end{figure}

The artifact issue becomes evident in \cref{fig:feng_basis_comp_mock}, which displays synthetic data generated through stochastic gradient descent within an isotropic quadratic potential coupled with isotropic noise. With 2,500 dimensions, the model mirrors the scale of a layer in the fully connected neural network that Feng \& Tu investigated. The weight series comprises 12,000 steps, which correspond to ten epochs of training this network. Analyzing this data with the weight covariance eigenbasis seemingly suggests anisotropic variance and correlation time. However, if the data is inspected without any basis change, both the variance and correlation time appear isotropically distributed as anticipated.

\begin{figure}[ht]
    \centering
    \includegraphics{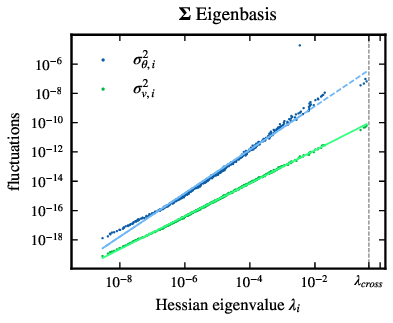}
    \includegraphics{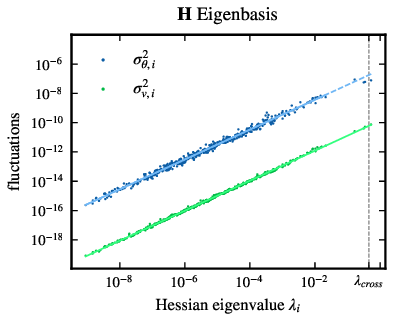}
    \\
    \includegraphics[trim=-0.083in 0in 0in 0in]{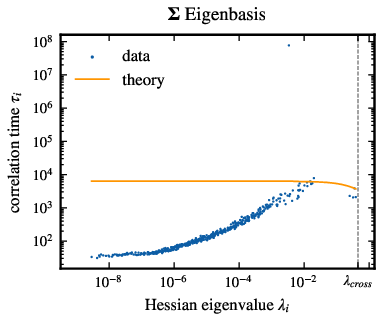}
    \includegraphics[trim=-0.083in 0in 0in 0in]{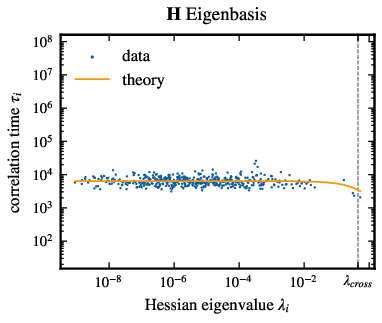}
    \caption{Comparison of weight and velocity variances for all 450 weights of the first convolutional layer of the LeNet, as discussed in the main text, analyzed in two different bases. In order to facilitate a more directly comparable analysis to f Feng \& Tu~\cite{Feng.2021}, the network was trained without weight decay for this specific analysis and the analysis period was limited to 10 epochs, as opposed to the usual 20 epochs. The columns represent different bases: for the left column $\Vec{p_i}$ are the eigenvectors of $\Vec{\Sigma}$ and for the right column $\Vec{p_i}$ are the eigenvectors of $\Vec{H}$. The mean velocity was subtracted in the right column. The rows illustrate the weight and velocity variance, $\sigma_{\theta, i}^2 = \Vec{p_i}^\top\Vec{\Sigma}\Vec{p_i}$, $\sigma_{v, i}^2 = \Vec{p_i}^\top\Vec{\Sigma_v}\Vec{p_i}$ (top row), and the correlation time $\tau_i = 2 \sigma_{\theta, i}^2 / \sigma_{v, i}^2$ (bottom row). The second derivative of the corresponding direction is depicted on the x-axis, $\lambda_i = \Vec{p_i}^\top\Vec{H}\Vec{p_i}$. The top row solid lines indicate fit regions for a linear fit. For the $\Vec{H}$ eigenbasis, the respective exponent of the power law relation is $1.018\pm 0.008$ for weight variance and $1.017\pm 0.002$ for velocity variance with a $2\sigma$-error. For the $\Vec{\Sigma}$ eigenbasis, the corresponding exponent is $1.537\pm 0.012$ for weight variance and $1.134\pm 0.002$ for velocity variance.
    }
    \label{fig:feng_basis_comp_real}
\end{figure}

To navigate around this key issue associated with the eigenbasis of $\Vec{\Sigma}$, we adopted the eigenvectors of the Hessian matrix. Unlike $\Vec{\Sigma}$, the Hessian is not computed as an average over update steps but can, in theory, be precisely calculated for any given weight vector. Consequently, the Hessian matrix does not suffer from finite size effects. The difference between these two bases for actual data is visible in \cref{fig:feng_basis_comp_real}. Here, we analyzed only the weights of the first convolutional layer of the LeNet from the main text to ensure comparability with Feng \& Tu's results. In this specific comparison, the network was trained without weight decay. Due to this and the fact that we are only investigating the weights of one layer, $\lambda_{\textrm{cross}}$ is significantly larger than all Hessian eigenvalues. As a result, when analyzing in the eigenbasis of the Hessian matrix related to this layer, both the variance and the correlation time align well with the prediction for smaller Hessian eigenvalues.

However, analyzing in the eigenbasis of the weight covariance matrix, the correlation time appears heavily dependent on the second derivative of the loss in the given direction. Additionally, the relationship between the weight variance and the second derivative shifts and more closely aligns with Feng \& Tu's results, with the exponent of a power law fit being significantly larger than one. In this regime, our theory predicts proportionality between weights and Hessian eigenvalues, reflected by an exponent of one for a power law fit. The first principal component, which Feng \& Tu referred to as the drift mode, stands out due to its unusually long correlation time. This is to be expected, as this is the direction in which the weights are moving at an approximately constant velocity.

\FloatBarrier
\subsection{Drawing with replacement}
\label{sec:appendix_with_replacement}

To confirm that the results obtained are indeed affected by the correlations present in SGD noise, due to the epoch-based learning strategy, we reapply the analysis described in the main text. In this instance, however, we deviate from our previous method of choosing examples for each batch within an epoch without replacement. Instead, we select examples with replacement from the complete pool of examples for every batch. This modification during the analysis period allows a more complete assessment of the impact of correlations  on the derived results.

\begin{figure}[ht]
    \centering
    \includegraphics[]{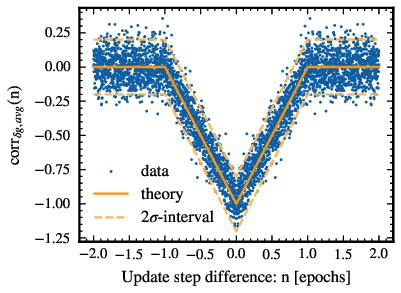}
    \includegraphics[]{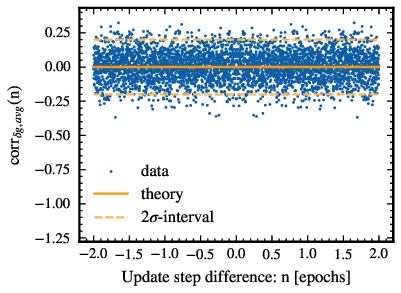}
    \caption{Autocorrelations of the SGD noise compared for drawing examples without replacement (left) and with replacement (right).}
    \label{fig:comp_wr_correlation}
\end{figure}

\cref{fig:comp_wr_correlation} offers clear visual proof that when examples are selected with replacement, the previously noted anti-correlations within the SGD noise vanish. This observation confirms our hypothesis that the anti-correlations mentioned in the main text are indeed an outcome of the epoch-based learning technique. Consequently, we can predict that this change will influence the behavior of the weight and velocity variance. As previously discussed, the theoretical results we have achieved for Hessian eigenvectors with eigenvalues exceeding $\lambda_{\textrm{cross}}$ conform to what one would predict in the absence of any correlation within the noise. Therefore, when examples are drawn with replacement, we anticipate the weight variance to be isotropic in all directions, while the velocity variance should remain unchanged.

\begin{figure}[ht]
    \centering
    \includegraphics{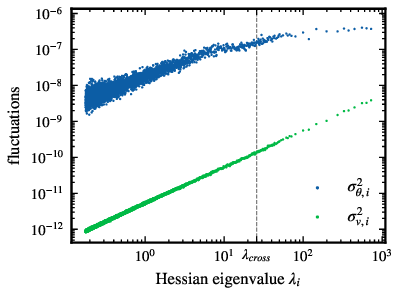}
    \includegraphics{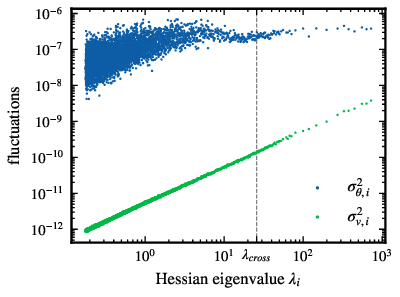}
    \\
    \includegraphics[trim=-0.083in 0in 0in 0in]{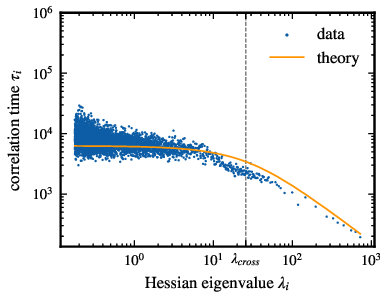}
    \includegraphics[trim=-0.083in 0in 0in 0in]{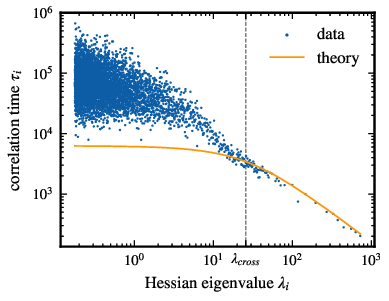}
    \caption{Relationship between Hessian eigenvalues and the variances of weights and velocities, as well as correlation times. For the left column the examples are drawn  in epochs without replacement and for the right column the examples are drawn with replacement.}
    \label{fig:comp_wr}
\end{figure}

 Upon reviewing \cref{fig:comp_wr}, it is clear that the velocity variance stays unchanged as predicted. However, while the weight variance remains constant for a broader subset of Hessian eigenvalues, it reduces for extremely small eigenvalues. Likewise, the correlation time is still limited for these minuscule Hessian eigenvalues. These deviations can be attributed to the finite time frame of the analysis period, comprising 20,000 update steps. This limited time window sets a cap on the maximum correlation time, consequently leading to a decreased weight variance for these small Hessian eigenvalues. Despite this, it is noteworthy that this maximum correlation time is still roughly one order of magnitude longer than the maximum correlation time induced by the correlations arising from the epoch-based learning approach.

%%%%%%%%%%%%%%%%%%%%%%%%%%%%%%%%%%%%%%%%%%%%%%%%%%
%%%%%%%%%%%%%%%%%%%%%%%%%%%%%%%%%%%%%%%%%%%%%%%%%%
%%%%%%%%%%%%%%%%%%%%%%%%%%%%%%%%%%%%%%%%%%%%%%%%%%
\FloatBarrier
\subsection{Weight variance for very small Hessian eigenvalues}
\label{sec:appendix_hessian_very_small}

As discussed in \cref{sec:var_main}, under the assumption that the gradient noise covariance is proportional to the Hessian, the stationary weight variances along Hessian eigendirections are expected to scale linearly with the corresponding eigenvalues, i.e., $\sigma_{\theta, i}^2 \propto \lambda_i$ for $\lambda_i<\lambda_{\rm cross}$. This prediction holds well across a broad range of eigenvalues in our experiments.

\begin{figure}[ht]
    \centering
    \includegraphics{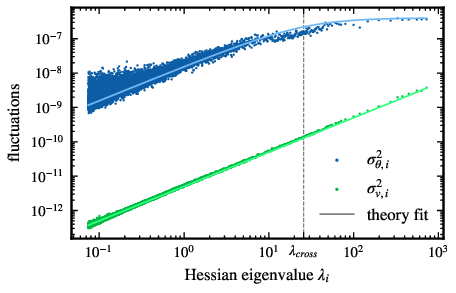}
    \caption{Relationship between Hessian eigenvalues and the variances of weights and velocities. The solid lines signify theoretical predictions from \cref{variances.eq}, assuming $\vec{C} \approx c_0 \vec{H}$ with $c_0$ fitted to the data. The plot extends the analysis to the top 10,000 Hessian eigenvectors. While the predicted proportionality holds well over a broad range, slight deviations appear in the lowest-curvature directions, where some weight variances begin to saturate and no longer decrease with $\lambda_i$. These deviations are consistent with either non-equilibrated weights in flat directions or corrections to the noise anti-correlations due to fluctuating weights.}
    \label{fig:appendix_weight_variance_10k}
\end{figure}

However, slight deviations from this linear relationship become apparent in the lowest-curvature directions. In particular, for a subset of the smallest eigenvalues among the top 5000, the observed weight variances slightly exceed the predicted values. These deviations become more pronounced when extending the analysis to the top 10,000 eigenvalues, as shown in \cref{fig:appendix_weight_variance_10k}.

This behavior may arise from several factors. First, modifications to the noise anti-correlations due to a fluctuating weight vector can influence the linear variance–curvature relationship, particularly for very small Hessian eigenvalues. In \cref{sec:appendix_non_static_weights}, we discussed an extended model that accounts for such fluctuations, predicting a minimal floor for the weight variance as $\lambda_i \to 0$.

Another possibility is that the weights have not yet fully equilibrated in these extremely flat directions within the training duration considered, leading to residual variances above the predicted values. The smallest eigenvalues observed are on the order of $10^{-1}$. To illustrate the plausibility of slow equilibration, consider full gradient descent with the final learning rate of about $10^{-5}$. Even with momentum $\beta = 0.9$, it would take roughly $10^5$ update steps to make substantial progress in such directions. In our setup, this corresponds to around 100 epochs -- while training lasted only 300 epochs in total. Using SGD instead of full-batch GD likely increases this timescale further due to additional noise.

Despite these deviations, the overall trend of variance–curvature proportionality remains robust across a wide range of eigenvalues. Where deviations do occur, the weight variance is still significantly lower than predictions from prior literature, which assumed a uniform variance across all directions equal to that in the sharpest mode.

%%%%%%%%%%%%%%%%%%%%%%%%%%%%%%%%%%%%%%%%%%%%%%%%%%
%%%%%%%%%%%%%%%%%%%%%%%%%%%%%%%%%%%%%%%%%%%%%%%%%%
%%%%%%%%%%%%%%%%%%%%%%%%%%%%%%%%%%%%%%%%%%%%%%%%%%
\FloatBarrier
\subsection{Testing different Hyperparameters}
\label{sec:appendix_hyperparameters}

In this section we examine the dependence of the theoretical predictions on the three hyperparameters learning rate $\eta$, momentum $\beta$ and batch size $S$. For this, we train the LeNet again for 300 epochs, using an exponential learning rate schedule that reduces the learning rate by a factor of $0.98$ every epoch and afterwards we perform the numerical analysis as described in the main text.

However, we now train the network several times, always varying one of the hyperparameters while keeping the other two fixed. If not varied, the momentum was set to 0.90 and the batch size was set to 64. To ensure that training always achieves a comparatively high training accuracy, the initial learning rate was set to 0.005 when the batch size is varied and to 0.02 when the momentum is varied. Five different values are examined for each hyperparameter. To investigate the dependencies on the learning rate, the values 0.005, 0.01, 0.02, 0.03, and 0.04 were used for training. For momentum, the values 0.00, 0.50, 0.75, 0.90, and 0.95 were examined, and for batch size, the values 32, 50, 64, 100, and 128 were examined.

In addition, the training was repeated for five different seeds for each hyperparameter combination in order to obtain reliable results. This results in a relatively high computational cost. To reduce this, for the analysis in this section the weight and velocity variances are examined only in the subspace of the 2,000 largest Hessian eigenvalues and associated eigenvectors.

\begin{figure}[htb]
    \centering
    \includegraphics[width=0.95\textwidth]{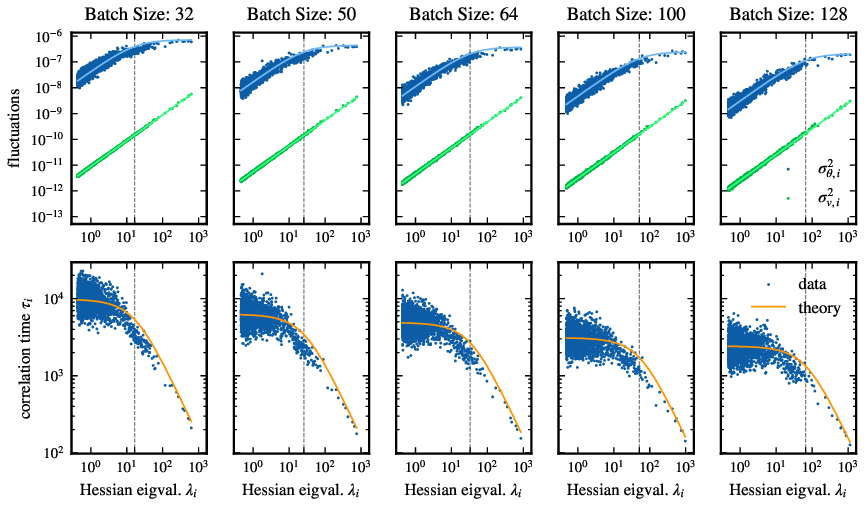}
    \caption{Testing the LeNet training with different hyperparameters. Here the relationship between the Hessian eigenvalues and the variances and correlation times for varying batch size is shown as an example. The momentum was set to 0.90 and the initial learning rate was set to 0.005.}
    \label{fig:appendix_batch_size_varied}
\end{figure}

\cref{fig:appendix_batch_size_varied} shows as an example the weight and velocity variances as well as the correlation times for different values of the batch size for one training seed each. It can be seen that the theory is not only valid for the hyperparameter combination from the previous section, but is also generally applicable for different hyperparameters. In particular, we see that there is still good agreement with the theory even if the strictly necessary condition for the theoretical derivation of the noise autocorrelation, that the number of batches per epoch $M = N/S$ is an integer, is not met.

To further examine the predictions of the theory for the hyperparametric dependencies, we now focus on the two quantities of the maximum correlation time $\tau_{\rm SGD}$ and the Hessian eigenvalue crossover value $\lambda_{\rm cross}$ and recall the theoretical predictions for these quantities: 
\begin{subequations}
\begin{align}
    \tau_{\textrm{SGD}} &= \frac{N}{3 S}\frac{1+\beta}{1-\beta} \ ,\\
    \lambda_{\textrm{cross}} &= \frac{3 S (1-\beta)}{\eta N} \ ,
\end{align}
\end{subequations}
where $N$ is the number of examples in the training data set.

For the evaluation of the dependence of these variables on the hyperparameters, they were determined as follows for the various hyperparameter combinations using the data from the respective correlation time plot. For the maximum correlation time $\tau_{\textrm{SGD}}$, the average of all correlation times was taken for which the corresponding Hessian eigenvalue is smaller than the theoretical crossover value. However, the result for the numerically determined maximum correlation time is similar when taking the average of all determined correlation times for a hyperparameter combination, since only very few Hessian eigenvalues are larger than the crossover value.

\begin{figure}[htb]
    \centering
    \includegraphics[width=0.95\textwidth]{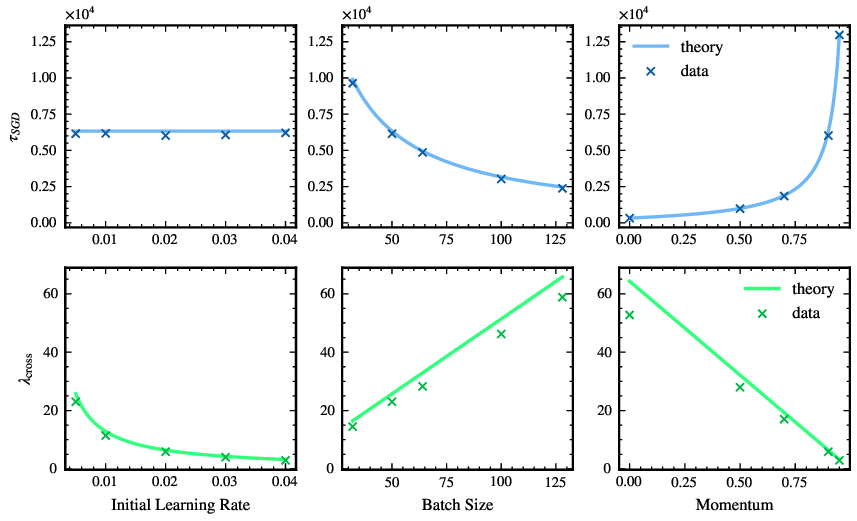}
    \caption{The empirically determined maximum correlation time, $\tau_{\textrm{SGD}}$, and the empirically determined crossover value $\lambda_{\rm cross}$ for the training of the LeNet with different hyperparameters, averaged over five different seeds for each set of hyperparameters. The predictions of our theory are shown as a solid line. The fluctuations between different seeds are smaller than the marker size and are therefore not included in the figure.}
    \label{fig:numerics_tau_lambda_vary}
\end{figure}

For the numerical determination of the crossover value $\lambda_{\rm cross}$, a linear function was first fitted to the correlation times of the 5 largest Hessian eigenvalues in the log-log plot of the correlation times against the Hessian eigenvalues. In this region of the first 5 values, the correlation time is always clearly dependent on the Hessian eigenvalue and does not yet belong to the region of constant correlation times. The intersection of the fitted line with the numerically determined maximum correlation time $\tau_{\textrm{SGD}}$ is then taken as the crossover value $\lambda_{\rm cross}$. If the numerically determined correlation times follow the theory exactly, then the correlation times determined in this way for $\tau_{\textrm{SGD}}$ and $\lambda_{\rm cross}$ would also follow the theory accurately.

And indeed, \cref{fig:numerics_tau_lambda_vary} shows a good agreement between the theory and the numerically determined values, although it should be noted that the deviations are larger than the random fluctuations between the different seeds.

%%%%%%%%%%%%%%%%%%%%%%%%%%%%%%%%%%%%%%%%%%%%%%%%%%
%%%%%%%%%%%%%%%%%%%%%%%%%%%%%%%%%%%%%%%%%%%%%%%%%%
%%%%%%%%%%%%%%%%%%%%%%%%%%%%%%%%%%%%%%%%%%%%%%%%%%
\FloatBarrier
\subsection{Autocorrelation of SGD noise at the beginning of training}
\label{sec:appendix_correlation_start}
\begin{figure}[ht]
    \centering
    \includegraphics[]{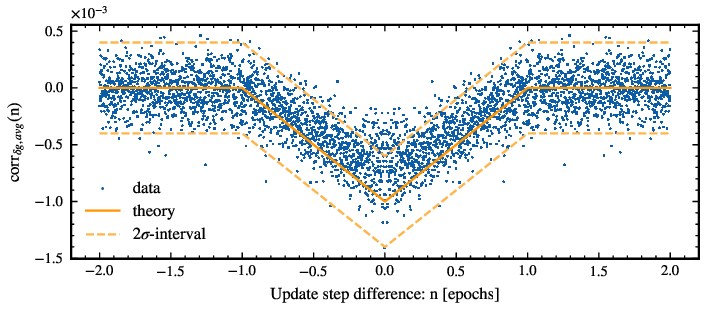}
    \caption{Autocorrelations of the SGD noise for the first ten epochs of the LeNet training schedule described in the main text. Since the Hessian matrix has a higher variability at this stage of training compared to later stages and its eigenvectors are not as suitable, for this experiment only, the noise terms are projected onto the first 2,500 principal components of the weight trajectory instead of the Hessian eigenvectors. As no variances are analyzed, no significant artifacts due to finite size effects are expected. The autocorrelation of the projected noise terms is computed individually and then averaged over the 2,500 principal components.}
    \label{fig:correlation_start}
\end{figure}

The theoretical prediction for the autocorrelation of SGD noise given by \cref{eq:noise_corr} was derived for a static weight vector and \cref{fig:noise_corr} shows strong agreement with empirical data from later stages of training. However, further numerical investigations reveal, that even at the beginning of training, when the network weights are clearly changing significantly, SGD noise still exhibits anti-correlations. \Cref{fig:correlation_start} demonstrates that these pronounced anti-correlations are also in good agreement with our theoretical prediction, although our analytic
theory relates only to a late phase of training.

%%%%%%%%%%%%%%%%%%%%%%%%%%%%%%%%%%%%%%%%%%%%%%%%%%
%%%%%%%%%%%%%%%%%%%%%%%%%%%%%%%%%%%%%%%%%%%%%%%%%%
%%%%%%%%%%%%%%%%%%%%%%%%%%%%%%%%%%%%%%%%%%%%%%%%%%
\FloatBarrier
\subsection{Relation between Hessian and noise covariance}
\label{sec:appendix_commutation_assumption}

The exact variance equation of \cref{thm:var_exact} described in \cref{sec:var_main} and the calculation shown in \cref{sec:appendix_var_calc} are still valid even if the previously mentioned assumption $[{\bf C},{\bf H}] \neq 0$ is not given. However, the calculated weight and velocity variances are no longer eigenvalues of the corresponding covariance matrices, but variances in the directions of the chosen eigenvector of the Hessian matrix.

Assuming that ${\bf C}$ and ${\bf H}$ do not necessarily commute, we can still project the update equations onto an arbitrary Hessian eigenvector $\vec{p}_i$ with eigenvalue $\lambda_i$, which gives us
\begin{subequations}
\begin{align}
    v_{k,i} &= -\eta\lambda_i\theta_{k-1,i} + \beta v_{k,i}  - \eta \vec{p} _i \cdot \vec{\delta g}_k \ ,\\
    \theta_{k,i} &= (1-\eta\lambda_i)\theta_{k-1,i} + \beta v_{k,i}  - \eta \vec{p} _i \cdot \vec{\delta g}_k \ .
\end{align}
\end{subequations}

Since we assume that $\vec{p}_i \cdot \vec{\delta g}_k$ is independent of weights and velocity, these two equations are decoupled for each individual Hessian eigenvector. As $\vec{p}_i \cdot \vec{\delta g}_k$ still follows the proposed anti-correlation, the calculations of \cref{sec:appendix_var_calc} can be performed similarly. Therefore, in a case without commutativity, the theory makes predictions for the variances along the Hessian eigenvectors, $\sigma_{\vec{\theta}, i}^2 \coloneqq \vec{p}_i^\top \vec{\Sigma}\vec{p}_i$ and $\sigma_{\vec{v}, i}^2 \coloneqq \vec{p}_i^\top \vec{\Sigma_v}\vec{p}_i$, depending on the noise variance in the given direction, $\sigma_{\vec{\delta g}, i}^2 \coloneqq \vec{p}_i^\top \vec{C}\vec{p}_i$, with $\sigma_{\vec{\delta g}, i}^2 = \left\langle\left(\vec{p}_i \cdot \vec{\delta g}_k\right)^2\right\rangle$. 
Independent of whether ${\bf C}$ and ${\bf H}$ commute, the results imply that the weight covariance, when restricted to the Hessian eigenspace corresponding to eigenvalues smaller than the crossover value, $\lambda_i<\lambda_{\rm cross}$, denoted by $\vec{\Sigma}_<$, is reduced compared to a setup without anti-correlations. To illustrate this effect, we examine the trace of the restricted weight covariance, ${\rm Tr}(\vec{\Sigma}_<)$. The theoretical prediction for its relative reduction is given by
\begin{align}
    \label{eq:appendix_var_reduction}
    \frac{{\rm Tr}(\vec{\Sigma}_{<,\textrm{with anti-correlations}})}{{\rm Tr}(\vec{\Sigma}_{<,\textrm{without anti-correlations}})} \approx \frac{\frac{M\eta}{3(1-\beta)} \sum_{i=i_{\rm cross}}^d \lambda_i \sigma_{\vec{\delta g}, i}^2}{\sum_{i=i_{\rm cross}}^d \sigma_{\vec{\delta g}, i}^2} \ ,
\end{align}
where $i_{\rm cross}$ is the index of the largest Hessian eigenvalue $\lambda_i$ that is still smaller than the crossover value $\lambda_{\rm cross}$. 
To develop an intuitive understanding of this result, consider a simplified case where all Hessian eigenvalues satisfying $\lambda_i<\lambda_{\rm cross}$ are assumed to be equal, such that $\lambda_i=\lambda_{\rm small}$. Under this assumption, the reduction formula \cref{eq:appendix_var_reduction} simplifies to $\frac{M\eta \lambda_{\rm small}}{3(1-\beta)}$. Since $\lambda_{\rm small}<\lambda_{\rm cross} \coloneqq \frac{3 (1-\beta)}{\eta M}$, this expression is necessarily smaller than one. This confirms that the presence of anti-correlations leads to a systematic reduction in the weight covariance without assuming that ${\bf C}$ and ${\bf H}$ commute.

Furthermore, the theory prediction for the correlation time $\tau_i$, defined as the ratio between the weight and the velocity variance in the eigendirection $\vec{p}_i$ of the Hessian, is also valid independently of the commutation relation between ${\bf C}$ and ${\bf H}$. However, the weight and the velocity variance in the eigendirection $\vec{p}_i$ defined above are no longer eigenvalues of the covariance matrices $\bf \Sigma$ and $\bf \Sigma_v$ if $\vec{p}_i$ is not also an eigenvector of ${\bf C}$. Nevertheless, these variances are still a good approximation for the actual eigenvalues, as long as ${\bf C}$ and ${\bf H}$ commute approximately.

\begin{figure}[tbh]
    \centering
    \includegraphics{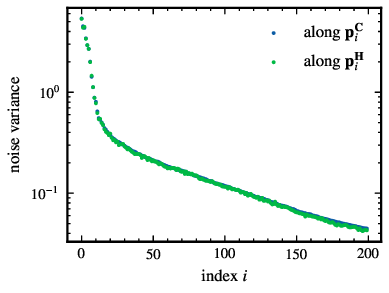}
    \caption{Noise variance during the analysis period of the LeNet considered in the main text. We analyzed the gradient noise recorded in the subspace of the 5,000 largest Hessian eigenvalues and corresponding eigenvectors ${\bf p}_i^{\bf H}$. We calculated the covariance matrix of the recorded noise, ${\bf C}$, and its eigenvectors ${\bf p}_i^{\bf C}$ in descending order according to their eigenvalues. We then plotted the largest eigenvalues of the noise covariance, which is the noise variance along ${\bf p}_i^{\bf C}$, together with the noise variance along ${\bf p}_i^{\bf H}$. When ${\bf C}$ and ${\bf H}$ commute, there exists a simultaneous eigenbasis for both matrices, and the shown variances would align perfectly. We find that the variances show very good alignment, indicating that both matrices commute approximately.}
    \label{fig:appendix_H_and_C_eigenbasis}
\end{figure}

While empirical investigations show that ${\bf C}$ and ${\bf H}$ do not commute exactly, they show a high alignment and suggest that the Hessian eigenbasis is a good approximation for the eigenbasis of the noise covariance (see \cref{sec:related_works_hessian_noise}). We find such an approximate commutativity as well. We investigated the Hessian and the covariance of the recorded noise in the approximated eigenspace of the Hessian during the analysis period of the LeNet considered in the main text. In \cref{fig:appendix_H_and_C_eigenbasis} one can see, that the actual 200 largest eigenvalues of the noise covariance align well with the variances in the directions of the 200 Hessian eigenvectors corresponding to the largest Hessian eigenvalues, indicating that the Hessian eigenbasis is very similar to the eigenbasis of the noise covariance.

\begin{figure}[tbh]
    \centering
    \includegraphics{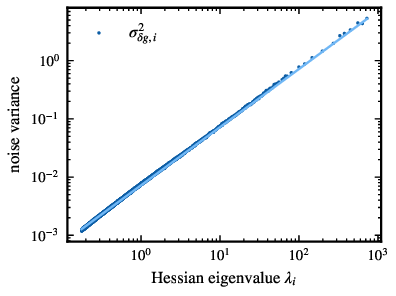}
    \caption{Relationship between Hessian eigenvalues and the variances of the recorded noise during the analysis period of the LeNet considered in the main text. The solid line signifies a linear fit. The exponent resulting from the power law relationship is $1.075\pm 0.002$ with a $2\sigma$-error. If the noise covariance were proportional to the Hessian matrix, the exponent should be equal to one.}
    \label{fig:appendix_H_and_C_linear}
\end{figure}

We also investigated how well the approximation of proportionality between ${\bf C}$ and ${\bf H}$ holds. For this we calculated the cosine similarity between both matrices in the approximated Hessian eigenspace, which is the normalized dot product between the flattened matrices. We found a cosine similarity of 0.82, which is similar to the empirical similarities found by Thomas et al.~\cite{Thomas.2020} and significantly higher than that of two random low-rank matrices. Additionally, we plotted the noise variances along the approximated Hessian eigenvectors against the corresponding Hessian eigenvalues and found that the two indeed follow an approximate linear relationship (see \cref{fig:appendix_H_and_C_linear}). So, while an exact proportionality between ${\bf C}$ and ${\bf H}$ is not satisfied, it seems to be a very good approximation.

%%%%%%%%%%%%%%%%%%%%%%%%%%%%%%%%%%%%%%%%%%%%%%%%%%
%%%%%%%%%%%%%%%%%%%%%%%%%%%%%%%%%%%%%%%%%%%%%%%%%%
%%%%%%%%%%%%%%%%%%%%%%%%%%%%%%%%%%%%%%%%%%%%%%%%%%
\FloatBarrier
\subsection{Different Network Architecture}
\label{sec:appendix_resnet20}

To further confirm our theoretical predictions within trained networks, in this section we turn to a more modern architecture. Instead of the previously used LeNet network, we examine the ResNet-20 network \cite{He.2016}. It is a convolutional network with significantly more convolutional layers than LeNet. It also uses residual blocks with residual connections, which allows for deeper network structures. As our loss function, we again employed Cross Entropy, along with an L2 regularization with a prefactor of $5\cdot10^{-3}$ and we did not use batch normalization. The number of layers, which is already indicated in the name with 20, is significantly higher than in the LeNet with just five layers. With approximately 272,000 parameters, the ResNet-20 also has significantly more parameters and the computational cost is significantly higher.

\begin{figure}[tbh]
    \centering
    \includegraphics[]{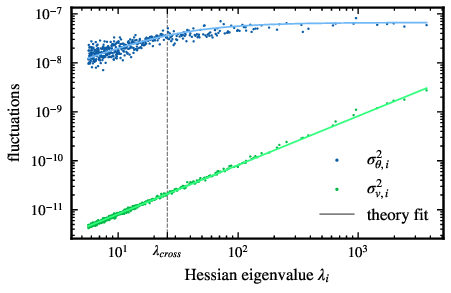}
    \includegraphics[]{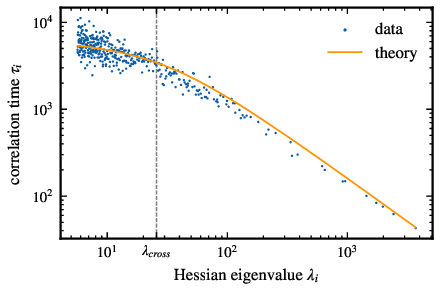}
    \caption{Relationship between Hessian eigenvalues and the variances and correlation times for a ResNet-20 trained on CIFAR10. In the left panel, we present the variances of weights and velocities. The solid lines signify theoretical predictions from \cref{variances.eq}, assuming $\vec{C} \approx c_0 \vec{H}$ with $c_0$ fitted to the data.  The right panel showcases the correlation time together  with  the  theoretical prediction resulting from \cref{variances.eq}, which does not require the $\vec{C} \approx c_0 \vec{H}$ assumption. The analysis was performed for the 400 largest Hessian eigenvalues and corresponding eigendirections.}
    \label{fig:appendix_resnet20_var}
\end{figure}

Therefore, in this section we limit ourselves to examining the weight and velocity variances in the subspace of the 400 largest Hessian eigenvalues and associated eigenvectors. The network was trained with SGD for 300 epochs using the same exponential learning rate schedule as before, with a learning rate of $5\cdot10^{-3}$, a momentum parameter of $0.9$, and a minibatch size of $50$. This setup achieves 100\% training accuracy and 76\% testing accuracy. In \cref{fig:appendix_resnet20_var} one can observe a good agreement between the theory predictions for the variances and correlation times and the numerical observations.

\FloatBarrier

\bibliography{bibliography}

\end{document}